\documentclass{article} 
\usepackage{iclr2027_conference,times}
\iclrfinalcopy

\usepackage{amsmath,amsfonts,bm}

\def\eqref#1{equation~\ref{#1}}

\def\1{\bm{1}}

\DeclareMathAlphabet{\mathsfit}{\encodingdefault}{\sfdefault}{m}{sl}
\SetMathAlphabet{\mathsfit}{bold}{\encodingdefault}{\sfdefault}{bx}{n}

\usepackage{hyperref}
\usepackage{url}
\usepackage[utf8]{inputenc} 
\usepackage[T1]{fontenc}    
\usepackage{hyperref}       
\usepackage{url}            
\usepackage{booktabs}       
\usepackage{amsfonts}       
\usepackage{nicefrac}       
\usepackage{microtype}      
\usepackage{xcolor}         
\usepackage{amsmath}
\usepackage{graphicx}
\usepackage{subcaption}
\usepackage{multirow}
\usepackage{longtable}
\usepackage{pdflscape}

\hypersetup{hidelinks}
\title{Deep Divide-and-Reduce in\\Symbolic Regression}

\author{
Yusong Deng$^{1,2}$,
Yanjie Li$^{2,3}$,
Xin Ning$^{1,2}$,
Lina Yu$^{2}$,
Liping Zhang$^{2}$,
Shu Wei$^{2}$,
Mingzhu Wan$^{2}$,\\
\textbf{
Min Wu$^{2}$,
Weijun Li$^{1,2,3}$
}
\\
$^{1}$School of Advanced Interdisciplinary Sciences, University of Chinese Academy of Sciences, Beijing, China\\
$^{2}$AnnLab, Institute of Semiconductors, Chinese Academy of Sciences, Beijing, China\\
$^{3}$Zhongguancun Academy, Beijing, China\\
}

\begin{document}

\maketitle
\lhead{}   

\begin{abstract}
Symbolic regression (SR) aims to discover underlying mathematical expressions from data while preserving interpretability. Most existing learning-based SR methods primarily optimize expressions from observations without explicitly exploiting their structural mathematical properties. AI Feynman introduced a complementary paradigm that leverages such properties to recursively decompose complex expressions, but its decomposition criteria cover only restricted structural forms and its treatment of nested composition can require brute-force search over candidate sub-expressions. Building on this paradigm, we propose Deep Divide-and-Reduce in Symbolic Regression (DDRSR), a mathematically grounded framework that systematically generalizes expression decomposition and variable reduction. DDRSR extends translational symmetry to coefficient- and exponent-interfered forms, enables variable separation under overlapping variables and additive constant offsets, and generalizes the identification of nested compositional structures. We further characterize an intrinsic non-identifiability limitation of decomposition when no effective variable separation is induced. Experiments across multiple symbolic regression algorithms and benchmark datasets show that DDRSR identifies a broader range of decomposable structures than AI Feynman and overall improves downstream regression accuracy and exact-expression recovery.
\end{abstract}

\section{Introduction}
Symbolic regression (SR) aims to discover interpretable analytical expressions from observational data and has become an important tool for scientific discovery. Existing SR methods are broadly dominated by search-based approaches, including genetic programming, reinforcement learning, and related heuristic search methods
\citep{petersen2019deep, sun2022symbolic, mundhenk2021symbolic, li2025gptmcts, liu2025camo, dong2025recent, al2024genetic, xu2024rsrm}, and pre-training-based approaches that formulate SR as sequence generation from large synthetic equation corpora \citep{biggio2021neural, kamienny2022end, vastl2024symformer, meidani2023snip, wu2023discovering, li2023transformer, shojaee2024llm}. Recent studies further combine learned priors with symbolic search to improve search efficiency and solution quality \citep{landajuela2022unified, xu2024rsrm, holt2023deep, liu2023snr, li2025gptmcts}. Despite their strong empirical performance, these approaches primarily optimize expressions from data and generally do not explicitly exploit the structural mathematical properties of the target function.

In contrast, AI Feynman \citep{udrescu2020aifeynman, udrescu2020aifeynman2} introduces a mathematically grounded paradigm that exploits translational symmetry, variable separability, and nested composition to recursively decompose complex expressions into simpler sub-problems. However, its decomposition criteria cover only relatively restricted structural forms. Its translational-symmetry rules are sensitive to coefficient or exponent interference, its separability criteria do not naturally handle overlapping variables or additive offsets, and its nested-composition mechanism may require searching over candidate sub-expressions. These restrictions substantially limit its applicability to complex, high-dimensional expressions. Concurrent work, NestyNet-SR \citep{xu2026nestynet}, also extends the AI Feynman paradigm, with an emphasis on structure discovery through analytic neural surrogates within an integrated SR pipeline.

To address these limitations, we propose Deep Divide-and-Reduce in Symbolic Regression (DDRSR), which systematically generalizes the decomposition principles of AI Feynman through mathematical proofs and derivations for broader variable composition and expression separation. Our primary innovations and contributions are summarized as follows:
\begin{itemize}

\item We extend translational symmetry, variable separability, and nested composition to substantially broader structural forms, including coefficient- and exponent-interfered variable compositions, overlapping-variable separability, additive constant offsets, and top-down nested separation.

\item We derive practical mathematical criteria for identifying these structures and establish a non-identifiability result showing the intrinsic limitation of decomposition without effective variable separation.

\item Across multiple SR solvers and benchmark datasets, DDRSR identifies more decomposable structures than AI Feynman and overall improves downstream regression and exact-expression recovery, with additional evaluations on runtime and robustness to noise.

\end{itemize}

\section{Related Work}
AI Feynman \citep{udrescu2020aifeynman, udrescu2020aifeynman2} exploits translational symmetry, variable separability, and nested composition to recursively reduce high-dimensional SR problems. However, its criteria mainly cover idealized separable structures: translational symmetry is sensitive to coefficient or exponent interference, conventional separability assumes disjoint variable groups and does not handle additive offsets, and nested-composition detection either requires unknown inner functions or candidate sub-expression search. These restrictions limit its applicability to overlapping-variable and structurally complex expressions. DDRSR generalizes these three decomposition mechanisms through analytical criteria that directly identify broader compositional structures. Detailed definitions of the AI Feynman criteria are provided in Appendix A.

Concurrent work, NestyNet-SR\citep{xu2026nestynet}, also extends the AI Feynman decomposition paradigm and uses derivative information from neural surrogates to identify richer structural patterns. Its main focus is on reliable structure discovery from sampled data within an integrated symbolic regression pipeline. In contrast, DDRSR focuses on the mathematical principles of decomposition itself, deriving explicit criteria and reconstruction mechanisms for generalized symmetry, separability, and nested structures, together with theoretical results on decomposition identifiability. Moreover, DDRSR serves as a model-agnostic decomposition framework compatible with different downstream SR methods.

\section{Methodology}
This section presents the primary definitions, criteria, and theorems utilized in our framework; detailed mathematical derivations and proofs are provided in Appendix \ref{Mathematical Derivations and Proofs}. Following the criterion-based reasoning adopted in AI Feynman, we derive characteristic analytical conditions that are necessarily satisfied by the corresponding structural forms and use them as practical criteria for structural identification. Unless explicitly stated as a theorem, these derivations are intended to provide analytical signatures for screening candidate structures rather than to establish complete necessary-and-sufficient characterizations.

In the tree representation of a mathematical expression, leaf nodes denote variables, whereas internal nodes represent mathematical operators. Within our methodology, the objective of translational symmetry is to consolidate multiple leaf nodes into a single equivalent leaf node, a process we define as bottom-up variable composition. Conversely, variable separability and nested composition attempt to partition the expression into two distinct sub-modules starting from the root node, a process we define as top-down expression separation.

Let \(\Omega\subseteq\mathbb{R}^n\) be a connected open domain and \(f:\Omega\to\mathbb{R}\). Unless otherwise specified, we assume that \(f\in C^3(\Omega)\), and that all denominators and transformed expressions involved in the criteria are well defined on the considered subdomain. 

\begin{figure}[thbp]
    \centering
    \includegraphics[width=\textwidth]{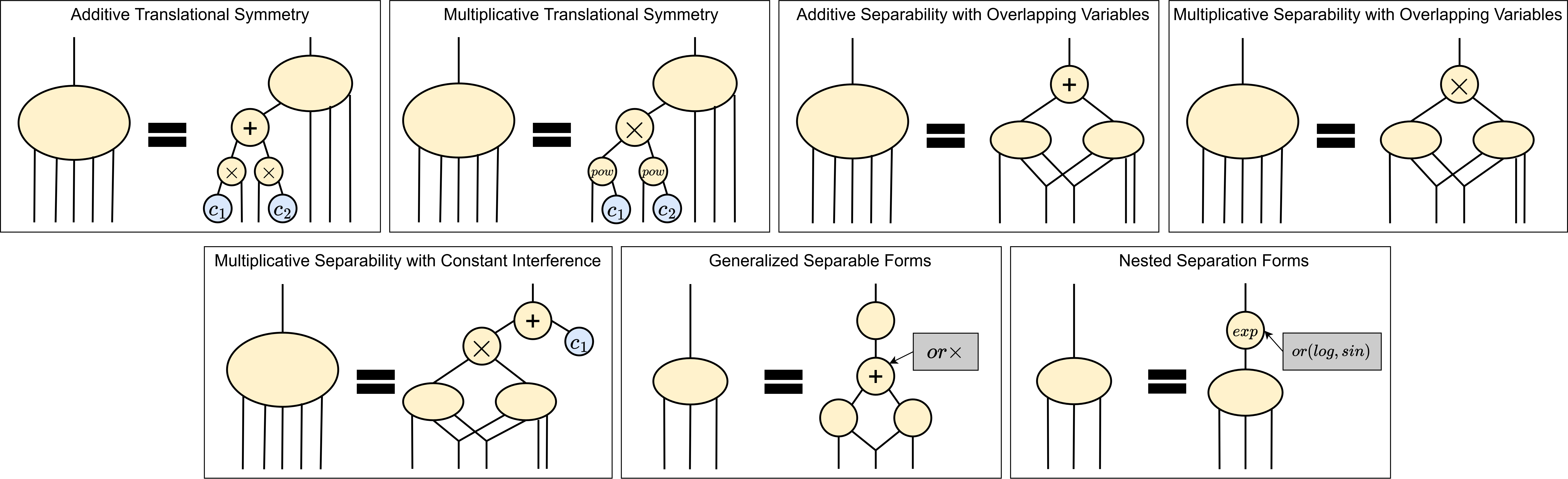}
    \caption{Examples of graph modularity that DDRSR can auto-discover. Lines denote real-valued variables and ovals denote functions.}
    \label{method}
\end{figure}

\subsection{Translational Symmetry}
\subsubsection{Additive Translational Symmetry}
\label{Additive Translational Symmetry}
\textbf{Definition}: For an $n$-dimensional function $f(X)$, if it can be formulated as $f(x_1, x_2, ...) = f(c_1 \times x_1 + c_2 \times x_2, ...)$, we state that $f(X)$ possesses additive translational symmetry with respect to the dimensions $x_1$ and $x_2$. 

\textbf{Criterion}: If, for any arbitrary $X$, the following condition holds:$$\frac{\partial f(X)}{\partial x_1} \div \frac{\partial f(X)}{\partial x_2} = c$$where $c$ is a constant, then $f(X)$ is deemed to possess additive translational symmetry across dimensions $x_1$ and $x_2$, yielding the relationship $\frac{c_1}{c_2} = c$.

\subsubsection{Multiplicative Translational Symmetry}
\textbf{Definition}: For an $n$-dimensional function $f(X)$, if it can be formulated as $f(x_1, x_2, ...) = f(x_1^{c_1} \times x_2^{c_2}, ...)$, we state that $f(X)$ possesses multiplicative translational symmetry with respect to the dimensions $x_1$ and $x_2$.

\textbf{Criterion}: If, for any arbitrary $X$, the following condition holds:$$\frac{\partial f(X)}{\partial x_1} \div \frac{\partial f(X)}{\partial x_2} \times \frac{x_1}{x_2} = c$$where $c$ is a constant, then $f(X)$ is deemed to possess multiplicative translational symmetry across dimensions $x_1$ and $x_2$, yielding the relationship $\frac{c_1}{c_2} = c$.  

\subsection{Variable Separability}
\subsubsection{Additive Separability with Overlapping Variables}
\textbf{Definition}: For an $n$-dimensional function $f(X)$, if it exhibits the form $f(SX_1, SX_2, SX_3) = g(SX_1, SX_2) + h(SX_2, SX_3)$, we state that $f(X)$ possesses additive separability with overlapping variables, specifically separable with respect to $SX_1$ and $SX_3$.

\textbf{Criterion}: If $f(X)$ satisfies the following structure for any $X$:
$$
f(SX_1, SX_2, SX_3) = f(SX_1, SX_2, SC_3) + f(SC_1, SX_2, SX_3) - f(SC_1, SX_2, SC_3)
$$
where $SC_i$ denotes setting the corresponding dimensions of $SX_i$ to constants, we determine that $f(X)$ possesses additive separability with respect to $SX_1$ and $SX_3$.

Furthermore, we can extract the sub-functions as follows:
$$
g(SX_1, SX_2) = f(SX_1, SX_2, SC_3) - f(SC_1, SX_2, SC_3)$$$$h(SX_2, SX_3) = f(SC_1, SX_2, SX_3)
$$

\subsubsection{Multiplicative Separability with Overlapping Variables}
\textbf{Definition}: For an $n$-dimensional function $f(X)$, if it exhibits the form $f(SX_1, SX_2, SX_3) = g(SX_1, SX_2) \times h(SX_2, SX_3)$, we state that $f(X)$ possesses multiplicative separability with overlapping variables, specifically separable with respect to $SX_1$ and $SX_3$.  

\textbf{Criterion}: If $f(X)$ satisfies the following structure for any $X$:
$$
f(SX_1, SX_2, SX_3) = \frac{f(SX_1, SX_2, SC_3) \times f(SC_1, SX_2, SX_3)}{f(SC_1, SX_2, SC_3)}
$$
we determine that $f(X)$ possesses multiplicative separability with respect to $SX_1$ and $SX_3$.

Furthermore, we can extract the sub-functions as follows:
$$
g(SX_1, SX_2) = \frac{f(SX_1, SX_2, SC_3)}{f(SC_1, SX_2, SC_3)}$$$$h(SX_2, SX_3) = f(SC_1, SX_2, SX_3)
$$

\subsubsection{Compositional Form of Additive Separability}
\textbf{Theorem 1}: For an $n$-dimensional function $f(X)$, if $f$ is additively separable with respect to dimensions $SX_1$ and $SX_3$, and concurrently additively separable with respect to $SX_2$ and $SX_3$, then $f$ can be rigorously expressed as:
$$
f(SX_1, SX_2, SX_3) = g(SX_1, SX_2) + h(SX_3)
$$

\subsubsection{Compositional Form of Multiplicative Separability}
\textbf{Theorem 2}: For an $n$-dimensional function $f(X)$, if $f$ is multiplicatively separable with respect to dimensions $SX_1$ and $SX_3$, and concurrently multiplicatively separable with respect to $SX_2$ and $SX_3$, then $f$ can be rigorously expressed as:  
$$
f(SX_1, SX_2, SX_3) = g(SX_1, SX_2) \times h(SX_3)
$$

\subsubsection{Multiplicative Separability with Additive Constant Offset}
This criterion is applied after standard additive and multiplicative separability have been excluded. In addition, we require \(\frac{\partial^2 f(X)}{\partial x_i\partial x_j}\neq 0\) on the considered subdomain to avoid degenerate cases.

\textbf{Definition}: For an $n$-dimensional function $f(X)$, if it takes the form $f(X) = g(x_i, SX) \times h(x_j, SX) + c$, where $SX$ denotes the remaining dimensions excluding $x_i$ and $x_j$, we state that $f(X)$ possesses multiplicative separability with a constant term across dimensions $x_i$ and $x_j$.

\textbf{Criterion}: Let $u_i(X) = \frac{\partial^2 f(X)}{\partial x_i\partial x_j} \div \frac{\partial f(X)}{\partial x_i}$ and $u_j(X) = \frac{\partial^2 f(X)}{\partial x_i\partial x_j} \div \frac{\partial f(X)}{\partial x_j}$. If $\frac{\partial u_i(X)}{\partial x_i} = 0$ and $\frac{\partial u_j(X)}{\partial x_j} = 0$ hold for any arbitrary $X$, we determine that \(f(X)\) may possess multiplicative separability with a constant term across dimensions \(x_i\) and \(x_j\). Furthermore, if the recovered constant \(c\) remains invariant for arbitrary values of the remaining variables, we confirm that \(f(X)\) indeed possesses this property.

\subsubsection{Separation Mechanism for Multiplicative Forms with Additive Constant Offset}
When \(f(X)\) satisfies the proposed differential criterion for multiplicative separability with a constant term, we further recover and verify the residual term.

\textbf{Theorem 3}: Given that $f(X)$ exhibits the form $f(X) = g(x_i, SX) \times h(x_j, SX) + c(SX)$, where $SX$ denotes the remaining dimensions excluding $x_i$ and $x_j$ and $c(SX)$ denotes a residual term that is independent of $x_i$ and $x_
j$. If two sample points are obtained for dimensions $x_i$ and $x_j$ respectively, denoted as $(x_{i1}, x_{i2})$ and $(x_{j1}, x_{j2})$, the constant term $c(SX)$ can be analytically formulated as:
$$
c(SX) = f(x_{i2}, x_{j2}, SX) - \frac{(f(x_{i2}, x_{j2}, SX) - f(x_{i2}, x_{j1}, SX)) \times (f(x_{i2}, x_{j2}, SX) - f(x_{i1}, x_{j2}, SX))}{f(x_{i2}, x_{j2}, SX) - f(x_{i1}, x_{j2}, SX) - f(x_{i2}, x_{j1}, SX) + f(x_{i1}, x_{j1}, SX)}
$$

\subsubsection{Generalized Separable Forms}
\label{Generalized Separable Forms Method}
\textbf{Definition}: For an $n$-dimensional function $f(X)$, if it takes the form $f(X) = U(g(x_i, SX) + h(x_j, SX))$ or $f(X) = U(g(x_i, SX) \times h(x_j, SX))$, where $U$ represents an arbitrary operator and $SX$ denotes the remaining dimensions excluding $x_i$ and $x_j$, we state that $f(X)$ possesses a generalized separable form.

\textbf{Criterion}: If the ratio $\frac{\partial f / \partial x_i}{\partial f / \partial x_j}$ exhibits multiplicative separability with respect to $x_i$ and $x_j$ for any arbitrary $X$, we determine that $f(X)$ possesses a generalized separable form across dimensions $x_i$ and $x_j$. 

However, under this criterion, it cannot be directly ascertained whether the intrinsic structure is additive or multiplicative. 

\subsection{Nested Composition}
\subsubsection{Nested Separation Form of Standard Logarithmic Functions}
\textbf{Definition}: For an $n$-dimensional function $f(X)$, if it exhibits the form $f(X) = \log(g(X))$, where $g(X)$ can undergo further separation, we state that $f(X)$ possesses a nested separation form of standard logarithmic functions.

\textbf{Criterion}: Let $f_{new}(X) = \exp(f(X))$. If $f_{new}(X)$ exhibits separability, we verify that $f(X)$ possesses a nested separation form of standard logarithmic functions.

Consequently, we reduce the problem to the separation of the non-nested function $f_{new}(X)$. Notably, functions formulated as $f(X) = \log(g(x_i) \times h(x_j))$ can be analytically rewritten as $f(X) = \log(g(x_i)) + \log(h(x_j))$. This structure intrinsically exhibits additive separability and will thus be correctly processed during the additive separation phase.  

\subsubsection{Nested Separation Form of Standard Exponential Functions}
\textbf{Definition}: For an $n$-dimensional function $f(X)$, if it exhibits the form $f(X) = \exp(g(X))$, where $g(X)$ can undergo further separation, we state that $f(X)$ possesses a nested separation form of standard exponential functions.

\textbf{Criterion}: Let $f_{new}(X) = \log(f(X))$. If $f_{new}(X)$ exhibits separability, we verify that $f(X)$ possesses a nested separation form of standard exponential functions.

Consequently, we reduce the problem to the separation of the non-nested function $f_{new}(X)$.  Notably, functions formulated as $f(X) = \exp(g(x_i) + h(x_j))$ can be analytically rewritten as $f(X) = \exp(g(x_i)) \times \exp(h(x_j))$. This structure intrinsically exhibits multiplicative separability and will thus be correctly processed during the multiplicative separation phase.  

\subsubsection{Nested Separation of Logarithmic Functions with Constant Terms}
Functions exhibiting the form $f(X) = c_1 \times \log(g(x_i, SX) \times h(x_j, SX)) + c_2$ can be mathematically rewritten as $f(X) = c_1 \times \log(g(x_i, SX)) + c_1 \times \log(h(x_j, SX)) + c_2$, allowing them to be systematically reduced to additive separation.

For functions exhibiting the form $f(X) = \log(g(x_i, SX) + h(x_j, SX)) + c$, applying an inverse function mapping yields $\exp(f(X)) = \exp(c) \times g(x_i, SX) + \exp(c) \times h(x_j, SX)$. Therefore, the original function can be reduced to the standard logarithmic function ($\log$) separation.

Functions exhibiting the form $f(X) = c \times \log(g(x_i, SX) + h(x_j, SX))$ are not subjected to standalone separation criteria.

\subsubsection{Nested Separation of Exponential Functions with Constant Terms}
Functions exhibiting the form $f(X) = c_1 \times \exp(g(x_i, SX) + h(x_j, SX)) + c_2$ can be mathematically rewritten as $f(X) = c_1 \times \exp(g(x_i, SX)) \times \exp(h(x_j, SX)) + c_2$, allowing them to be reduced to multiplicative separation containing a constant term.

For functions exhibiting the form $f(X) = c \times \exp(g(x_i, SX) \times h(x_j, SX))$, applying an inverse function mapping yields $\log(f(X)) = \log(c) + g(x_i, SX) \times h(x_j, SX)$. Therefore, the original function can be reduced to the standard exponential function ($\exp$) separation.

Functions exhibiting the form $f(X) = \exp(g(x_i, SX) \times h(x_j, SX)) + c$ are not subjected to standalone separation criteria.

\subsubsection{Nested Separation Form of Standard Trigonometric Functions}
\textbf{Definition}: For an $n$-dimensional function $f(X)$, if it takes the form $f(X) = \sin(\text{or } \cos)(g(x_i, SX) + h(x_j, SX) + c)$ or $f(X) = \sin(\text{or } \cos)(g(x_i, SX) \times h(x_j, SX) + c)$, where $SX$ denotes the remaining dimensions excluding $x_i$ and $x_j$, we state that $f(X)$ possesses a nested separation form of standard trigonometric functions.

\textbf{Criterion}: Let $T(X) = \frac{\partial f(X)}{\partial x_i} \times \frac{\partial f(X)}{\partial x_j} \div (1 - f^2(X))$. If, for any arbitrary $X$, $T(X)$ exhibits multiplicative separability with respect to dimensions $x_i$ and $x_j$, $f(X)$ is considered a candidate for a standard trigonometric nested separation form.  

\subsubsection{Non-identifiability of Decomposition Without Variable Separability}
During the process of expression decomposition, if an expression is partitioned into two components where the variables involved in the second component are merely a subset of those in the first component—meaning no distinct variable separation has occurred—such a separation strategy is fundamentally non-identifiable without additional constraints.

\textbf{Theorem 4}: For an $n$-dimensional function $f(X)$, if $f(X)$ can be formulated as $f(SX_1, SX_2) = g(SX_1, SX_2) \textcircled{+} h(SX_2)$, where $\textcircled{+}$ denotes an associative and commutative binary operation with an identity element and an inverse operation $\textcircled{-}$, there exist infinitely many valid functional configurations for $g$ and $h$.

Consequently, it is analytically intractable to achieve identifiable expression decomposition under similar unconstrained conditions.

\section{Experiments}

\begin{figure}[htbp]
    \centering
    \begin{subfigure}[b]{0.45\textwidth}
        \centering
        \includegraphics[width=\textwidth]{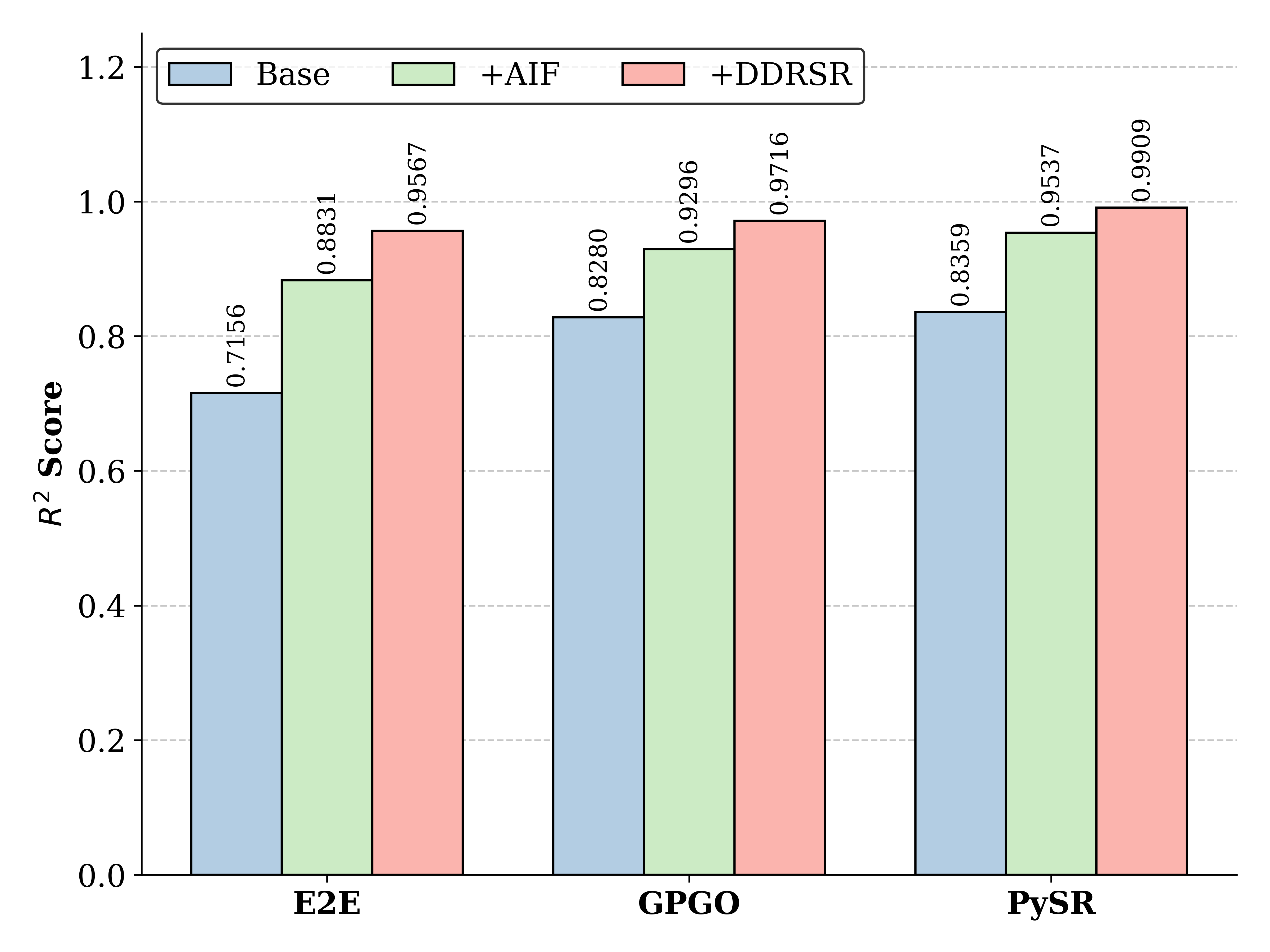}
        \caption{$R^2$ of Comprehensive Regression}
        \label{Comprehensive_Regression_R2}
    \end{subfigure}
    \hfill
    \begin{subfigure}[b]{0.45\textwidth}
        \centering
        \includegraphics[width=\textwidth]{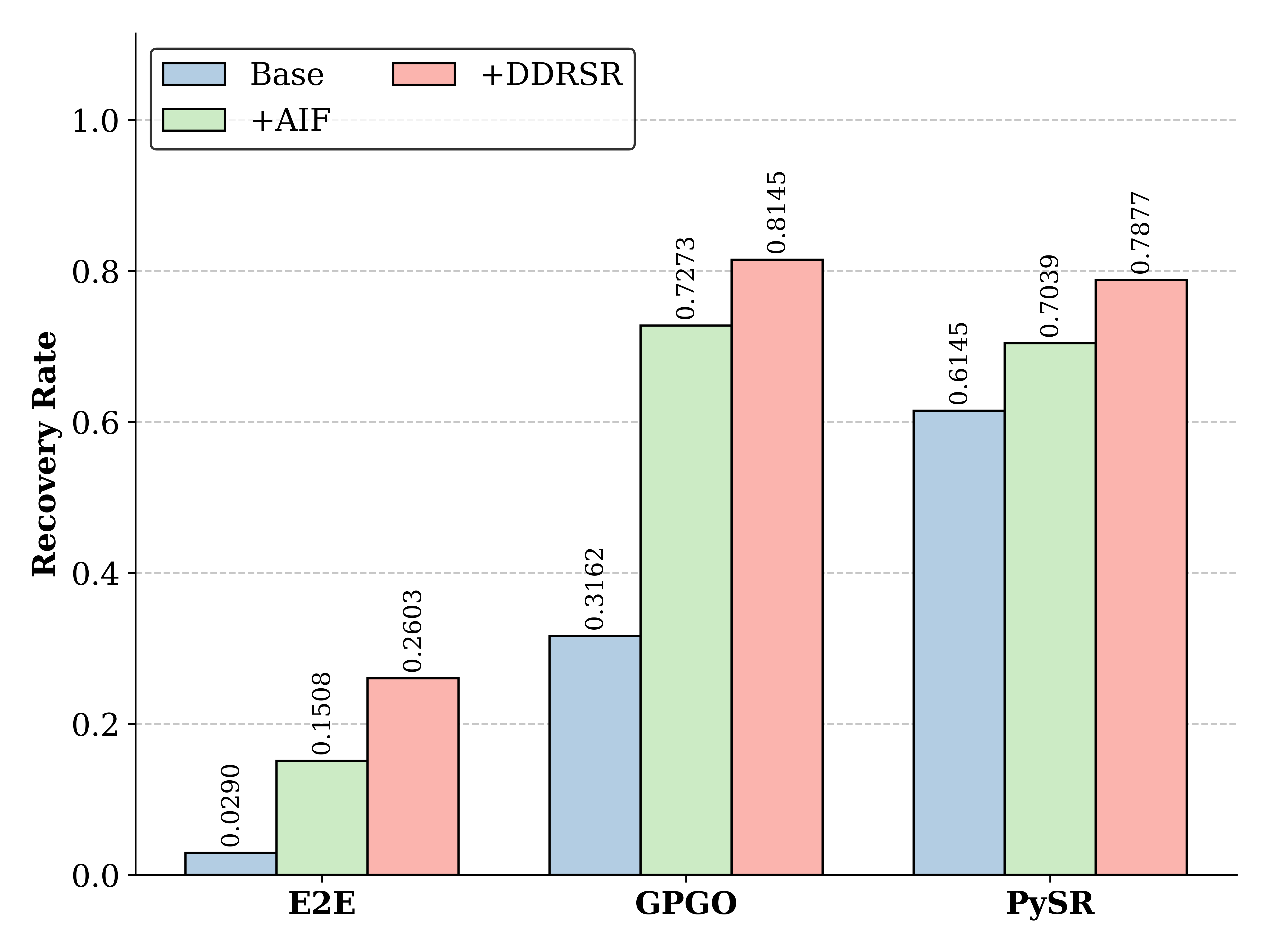}
        \caption{Recovery Rate of Comprehensive Regression}
        \label{Comprehensive_Regression_Rec}
    \end{subfigure}
    \vspace{0.3cm}
    \begin{subfigure}[b]{0.45\textwidth}
        \centering
        \includegraphics[width=\textwidth]{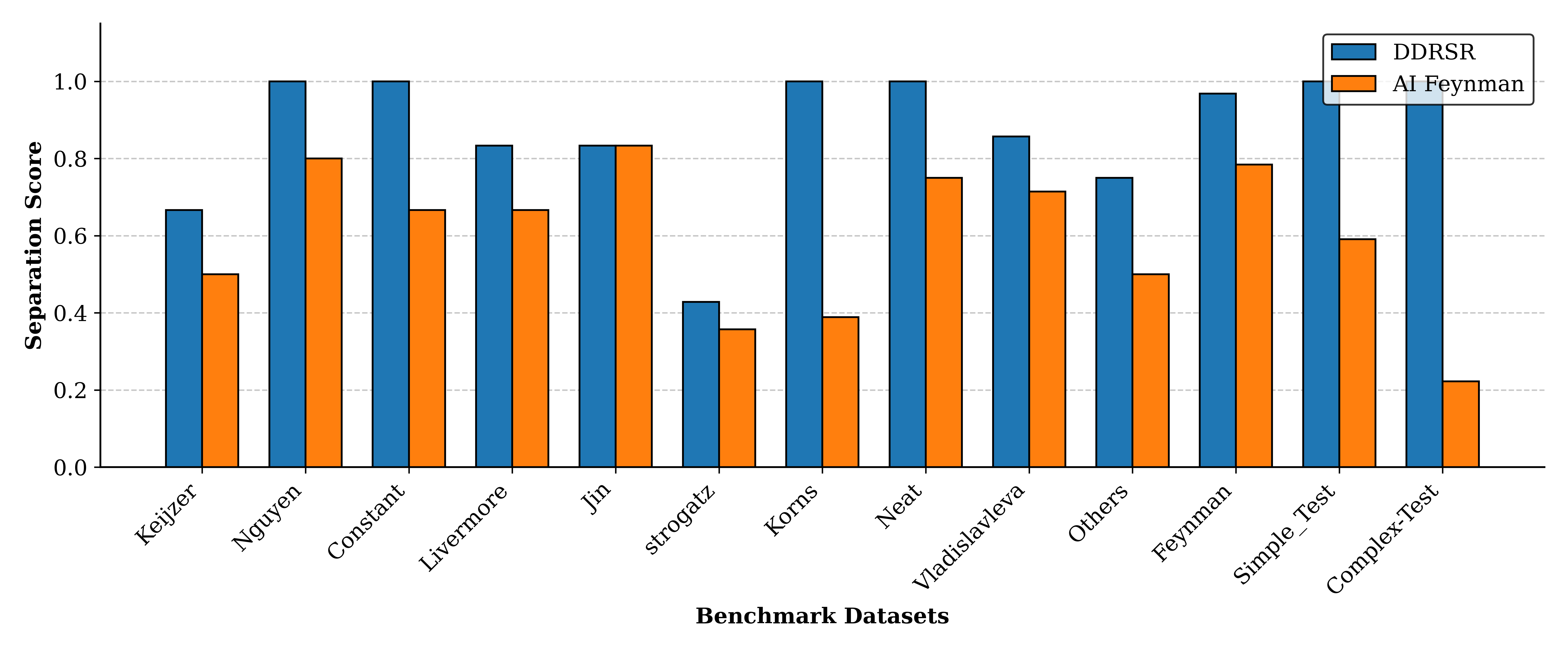}
        \caption{Relative Separation Score of Comprehensive Experiments}
        \label{Comprehensive_Separation}
    \end{subfigure}
    \hfill
    \begin{subfigure}[b]{0.45\textwidth}
        \centering
        \includegraphics[width=\textwidth]{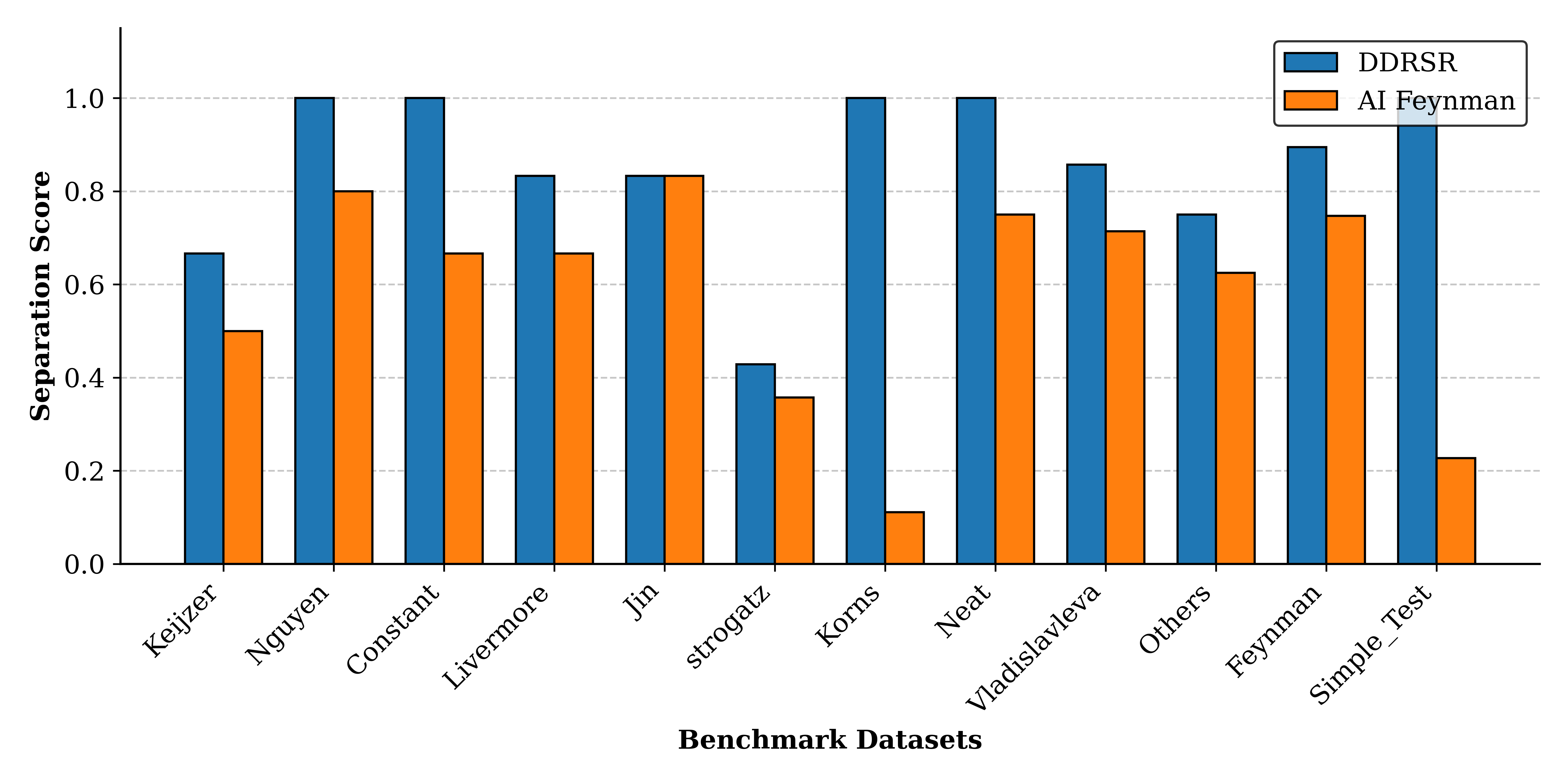}
        \caption{Relative Separation Score of Separation Ablation Experiments}
        \label{Separation_Separation}
    \end{subfigure}
    \vspace{0.3cm} 
    \begin{subfigure}[b]{0.45\textwidth}
        \centering
        \includegraphics[width=\textwidth]{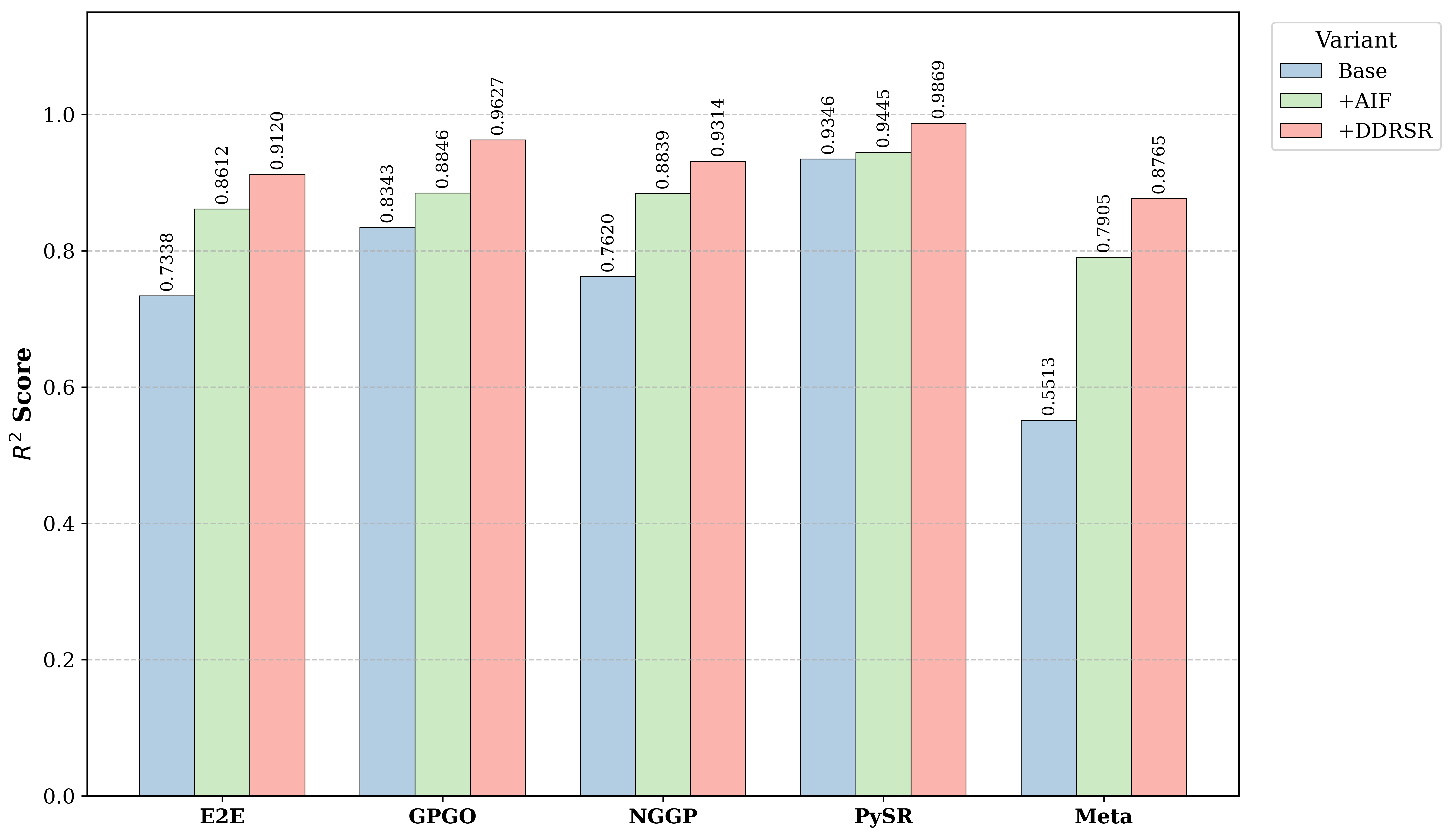}
        \caption{$R^2$ of Separation}
        \label{Separation_Regression}
    \end{subfigure}
    \hfill
    \begin{subfigure}[b]{0.45\textwidth}
        \centering
        \includegraphics[width=\textwidth]{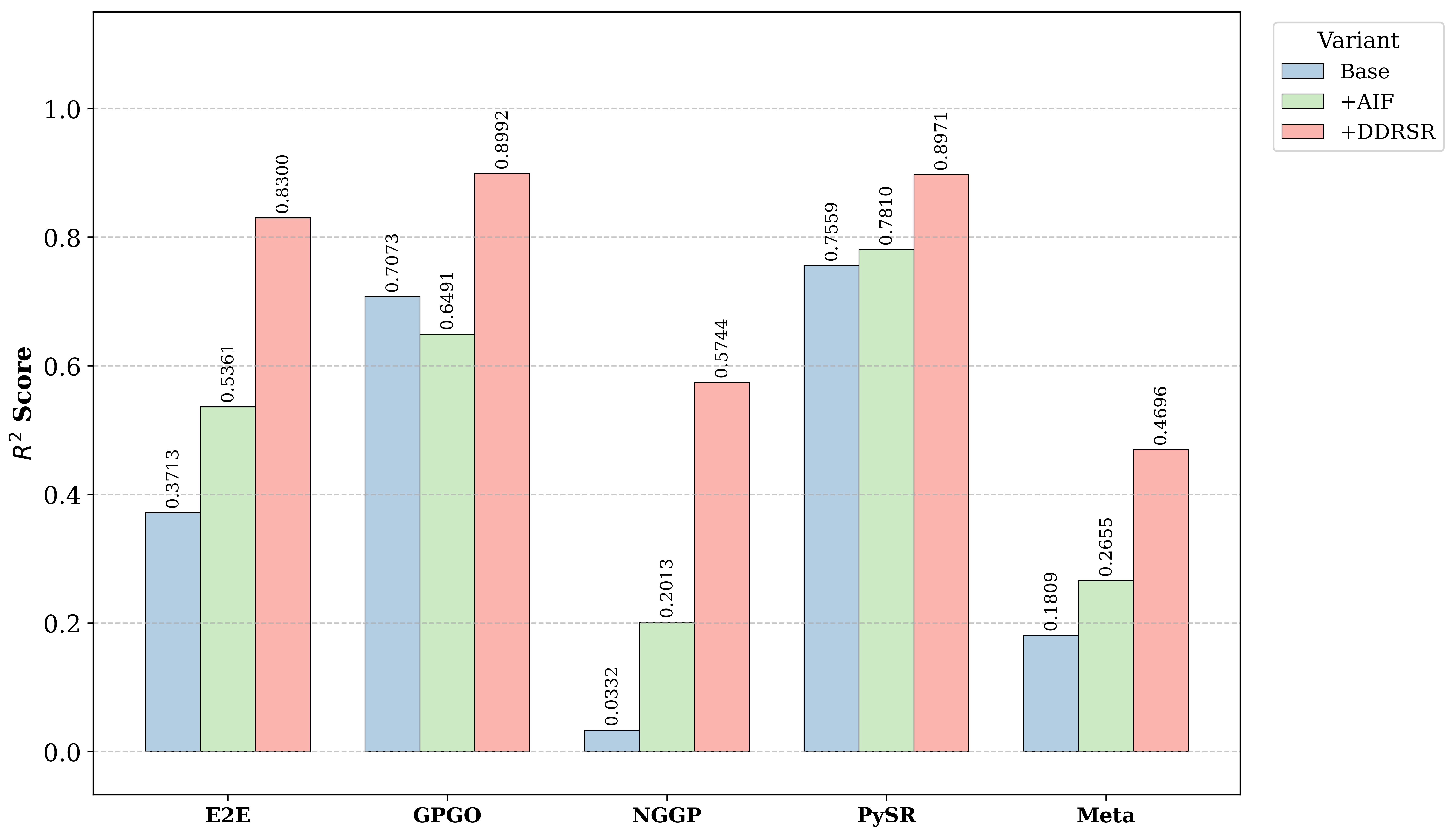}
        \caption{$R^2$ of Translation Symmetry}
        \label{Translation_Symmetry_Regression}
    \end{subfigure}

    \caption{Comprehensive and ablation evaluation of DDRSR against the undecomposed SR baseline and AI Feynman decomposition. (a) Mean $R^2$, (b) exact-expression recovery rate, and (c) Relative Separation Score in the comprehensive experiments. (d) Relative Separation Score and (e) mean $R^2$ for the top-down expression-separation ablation (variable separability and nested composition). (f) Mean $R^2$ for the bottom-up variable-composition ablation based on translational symmetry on complex-test.}
    \label{overall_results}
\end{figure}

\subsection{Experimental Setup and Procedure}

To evaluate the correctness and effectiveness of DDRSR, we apply the proposed decomposition principles to target expressions and subsequently solve the resulting sub-expressions using five symbolic regression algorithms: E2E \citep{kamienny2022end}, GP-GOMEA \citep{virgolin2021improving}, NGGP \citep{mundhenk2021symbolic}, PySR \citep{cranmer2023pysr}, and MetaSymNet \citep{li2023metasymnet}. We compare DDRSR with both the original SR methods without decomposition and the decomposition principles of AI Feynman, and additionally investigate robustness under noisy observations.

We use $R^2$ in the ablation and noise experiments, while the comprehensive experiments additionally report the recovery rate, defined as the fraction of runs whose predicted expressions are symbolically or mathematically equivalent to the ground truth. Each experiment is repeated five times, and the reported $R^2$ is averaged over the five runs. We further introduce a Relative Separation Score to quantify decomposition completeness: the score is 0 when no valid decomposition is identified, 0.5 when a valid but less complete decomposition is obtained than by another method, and 1 when the maximum number of correct decomposition patterns is identified.

To isolate the effect of the underlying decomposition principles, all brute-force-search-related components are disabled, since exhaustive enumeration may solve otherwise non-decomposable sub-expressions and thereby confound the comparison. In the idealized validation experiments, function values and required derivatives are evaluated directly from the analytical ground-truth functions to isolate and verify the proposed decomposition principles. To further assess DDRSR under non-ideal conditions, we explicitly introduce perturbations into the sampled observations, approximate derivatives using finite differences, and employ a relative-loss criterion with a numerical threshold to accommodate sampling and numerical errors. The resulting effects on decomposition reliability and downstream symbolic regression performance are systematically evaluated in the noise robustness experiments. Equality-based mathematical criteria are evaluated using a relative threshold of 0.01; the exact normalized loss definition, numerical stabilization term, and related implementation details are provided in Appendix \ref{Numerical Decision Criteria}.

For each unknown expression, DDRSR sequentially evaluates the proposed principles and recursively applies every successfully identified decomposition to the resulting sub-expressions until no branch can be further decomposed. Each terminal sub-expression is then processed once by the downstream SR algorithm, and the resulting expressions are recombined in reverse according to the decomposition structure to reconstruct the complete expression. Invalid numerical evaluations involving NaN or Inf are handled through finite resampling. Details of experimental procedure are in Appendix \ref{Numerical Decision Criteria}. Dataset formulas, SR hyperparameters, and hardware configurations are also provided in the Appendix.

\subsection{Comprehensive Symbolic Regression Experiments}
We evaluate the complete DDRSR framework, including variable separation, nested composition, and translational symmetry, on public benchmarks and custom-constructed test expressions. For these experiments, we select the three best-performing baseline SR methods, E2E, GP-GOMEA, and PySR, with the results summarized in Figure\ref{Comprehensive_Regression_R2} -- \ref{Comprehensive_Separation}.

DDRSR matches or exceeds AI Feynman in Relative Separation Score across all evaluated dataset groups (Figure \ref{Comprehensive_Separation}), confirming that the proposed principles can identify and mathematically parse a broader range of structural cases. In terms of $R^2$ (Figure \ref{Comprehensive_Regression_R2}), decomposition using either DDRSR or AI Feynman generally improves over the corresponding undecomposed SR baselines, demonstrating the benefit of reducing complex expressions before regression; dimension-wise statistics in Appendix further show that this advantage becomes increasingly pronounced as expression dimensionality and complexity increase. Recovery-rate results (Figure \ref{Comprehensive_Regression_Rec}), together with the $R^2$ results, show that DDRSR provides better overall performance on the evaluated benchmarks because it can identify and simplify structural cases that remain inaccessible to AI Feynman.

\subsection{Ablation Studies}
\textbf{Top-down expression separation.}

We independently evaluate top-down decomposition, consisting of variable separation and nested composition, on all datasets except complex-test (Figure \ref{Separation_Separation} and \ref{Separation_Regression}). DDRSR overall achieves higher Relative Separation Scores than AI Feynman, demonstrating its ability to recognize a broader range of separable structures. Both decomposition-based variants outperform the corresponding undecomposed SR methods in $R^2$, confirming that expression separation alone effectively reduces the difficulty of the target problem; DDRSR further outperforms AI Feynman because its generalized principles successfully separate a wider variety of structures.

\textbf{Bottom-up variable composition.}

We isolate the variable-composition mechanism, that based on translational symmetry, on the custom complex-test dataset (Figure \ref{Translation_Symmetry_Regression}). DDRSR achieves the best $R^2$ performance among the evaluated methods and obtains a Relative Separation Score of 0.7777, compared with 0.0555 for AI Feynman, demonstrating that the proposed variable-composition principles can identify and combine a substantially broader range of mathematical structures.

\subsection{Runtime Analysis}
We separately evaluate the running time of DDRSR, AI Feynman, and the baseline method without expression decomposition. The symbolic regression algorithms considered include E2E, PySR, and GP-GOMEA. The evaluation metric is the average running time of each expression over the entire test set, and the results are presented in Table \ref{table-time_test}.

\begin{table}[htbp]
\centering
\caption{Average running time (s).}
\label{table-time_test}
\begin{tabular}{c|ccc}
\hline
SR Method & Baseline & AI Feynman & DDRSR \\
\hline
E2E   & 4.31   & 17.84   & 30.06   \\
PySR  & 143.52 & 351.84  & 439.27  \\
GP-GOMEA  & 657.98 & 1246.43 & 1552.01 \\
\hline
\end{tabular}
\end{table}

DDRSR introduces additional computational overhead, which is expected. This is because DDRSR can achieve a more thorough decomposition of expressions, and a larger number of sub-expressions requires more frequent calls to symbolic regression algorithms, thereby resulting in increased running time.

\subsection{Robustness to Noise}
Strict mathematical principles are accompanied by exact analytical conditions, which are theoretically correct and reasonable under ideal experimental conditions. However, for non-ideal scenarios where data contain noise, the effectiveness of DDRSR requires further investigation. We add uniform random noise($\xi \sim \mathcal{U}(-\eta,\eta)$) to the data and count the number of expressions that fail to be successfully decomposed. The results are shown in Table \ref{table-noise_count}.

\begin{table}[htbp]
\centering
\caption{Number of expressions that fail to be decomposed under different noise levels.}
\label{table-noise_count}
\begin{tabular}{c|cccc}
\hline
 & $\eta$=0 & $\eta$=$10^{-7}$ & $\eta$=$10^{-6}$ & $\eta$=$10^{-5}$ \\
\hline
DDRSR & 23 & 37 & 66 & 113 \\
AI Feynman      & 41 & 52 & 81 & 141 \\
\hline
\end{tabular}
\end{table}

Theoretically, the decomposition principles of DDRSR involve more gradient computations, and gradients are more sensitive to noise. Therefore, noise is expected to have a greater impact on DDRSR than on AI Feynman. However, the experimental results show that, when using relative loss and thresholds for decision-making, DDRSR does not exhibit a substantially larger increase in decomposition failures than AI Feynman in the reported experiments. In addition, under noisy conditions, we apply AI Feynman and DDRSR to all expressions that remain decomposable, and perform symbolic regression on the resulting decomposed sub-expressions. We compare the experimental results under noisy conditions with those obtained without noise interference. Since AI Feynman and DDRSR retain different subsets of successfully decomposable expressions under noise, the PySR baseline in each row is evaluated on the corresponding method-specific subset. The results are shown in Table \ref{table-noise_regression}.

\begin{table}[htbp]
\centering
\caption{Regression performance under noise conditions.}
\label{table-noise_regression}
\begin{tabular}{c|ccc}
\hline
 & PySR\_R2 & Divide\_PySR\_R2 & Divide\_PySR\_R2\_Noise \\
\hline
AI Feynman & 0.8272 & 0.9814 & 0.9796 \\
DDRSR & 0.7730 & 0.9880 & 0.9861 \\
\hline
\end{tabular}
\end{table}

Table \ref{table-noise_regression} shows that once an expression is correctly decomposed by either DDRSR or AI Feynman, the improvement in symbolic regression performance brought by expression decomposition is almost identical to that under noise-free conditions.

Even when decomposition decisions fail due to noise interference, DDRSR falls back to applying the symbolic regression algorithm directly to the original expression. This fallback mechanism provides a practical safeguard against performance degradation caused by unsuccessful decomposition. Moreover, since the decomposition criteria require the corresponding mathematical properties to hold overall across the sampled domain, while noise is inherently random, noise-induced false-positive decompositions are unlikely to occur. Consequently, DDRSR is generally expected to perform no worse than direct symbolic regression in practice, while successful decomposition can provide additional performance gains, as also supported by our experimental results.

\section{Discussion and Conclusion}
To overcome the limited applicability of the decomposition principles in AI Feynman, we adopt a ``Divide-and-Reduce'' philosophy and systematically extend variable separation, nested composition, and translational symmetry, culminating in the proposed DDRSR. By identifying and mathematically decomposing a substantially broader range of structural patterns, DDRSR transforms complex symbolic regression problems into simpler sub-problems and thereby reduces the difficulty of downstream regression.

An important feature of this paradigm is its broad compatibility with existing symbolic regression algorithms: after decomposition, the resulting sub-expressions can be directly processed by different downstream SR methods and subsequently recombined to reconstruct the complete expression. Extensive experiments demonstrate that such decomposition overall improves downstream symbolic regression performance, with DDRSR achieving stronger predictive performance and recovery accuracy by recognizing and simplifying more complex structures than existing decomposition principles. Nevertheless, DDRSR and related mathematically grounded decomposition methods have inherent limitations, particularly in practical scenarios involving imperfect data and numerical estimation; these limitations and the corresponding directions for future development are discussed in Appendix \ref{Limitation and Future Work}.

\clearpage

\subsection*{AI use statement}

In this work, we used generative AI tools solely for English translation and language polishing of text originally written by the authors. Generative AI tools were not used for research ideation, methodology development, algorithm design, code generation, experimental design or implementation, data analysis, result interpretation, or the formulation of scientific claims and conclusions. All AI-assisted translations and language edits were carefully reviewed and verified by the authors to ensure that they accurately reflected the original content and intended meaning. We take full responsibility for the final content of this work, including all text, claims, and results.

\subsection*{Ethics statement}

This work does not involve human subjects, personal or sensitive data, or other activities that raise specific ethical concerns. To the best of our knowledge, this study does not introduce any notable ethical risks.

\subsection*{Reproducibility statement}

The principles and procedures underlying our method are precisely defined, enabling the proposed approach to be reproduced from the descriptions provided in this paper. The symbolic regression methods employed in our experiments are well-established and effective; although they involve stochastic optimization, repeated runs are expected to yield comparable overall results. The source code and implementation details will be made publicly available upon acceptance of this paper.



\bibliography{Reference.bib}

@article{sun2022symbolic,
  title={Symbolic physics learner: Discovering governing equations via monte carlo tree search},
  author={Sun, Fangzheng and Liu, Yang and Wang, Jian-Xun and Sun, Hao},
  journal={arXiv preprint arXiv:2205.13134},
  year={2022}
}

@article{petersen2019deep,
  title={Deep symbolic regression: Recovering mathematical expressions from data via risk-seeking policy gradients},
  author={Petersen, Brenden K and Landajuela, Mikel and Mundhenk, T Nathan and Santiago, Claudio P and Kim, Soo K and Kim, Joanne T},
  journal={arXiv preprint arXiv:1912.04871},
  year={2019}
}

@article{mundhenk2021symbolic,
  title={Symbolic regression via neural-guided genetic programming population seeding},
  author={Mundhenk, T Nathan and Landajuela, Mikel and Glatt, Ruben and Santiago, Claudio P and Faissol, Daniel M and Petersen, Brenden K},
  journal={arXiv preprint arXiv:2111.00053},
  year={2021}
}

@inproceedings{biggio2021neural,
  title     = {Neural Symbolic Regression that Scales},
  author    = {Biggio, Luca and Bendinelli, Tommaso and Neitz, Alexander and Lucchi, Aurelien and Parascandolo, Giambattista},
  booktitle = {Proceedings of the 38th International Conference on Machine Learning},
  pages     = {936--945},
  volume    = {139},
  series    = {Proceedings of Machine Learning Research},
  publisher = {PMLR},
  year      = {2021}
}

@article{kamienny2022end,
  title={End-to-end symbolic regression with transformers},
  author={Kamienny, Pierre-Alexandre and d'Ascoli, St{\'e}phane and Lample, Guillaume and Charton, Fran{\c{c}}ois},
  journal={Advances in Neural Information Processing Systems},
  volume={35},
  pages={10269--10281},
  year={2022}
}

@article{vastl2024symformer,
  title={Symformer: End-to-end symbolic regression using transformer-based architecture},
  author={Vastl, Martin and Kulh{\'a}nek, Jon{\'a}{\v{s}} and Kubal{\'\i}k, Ji{\v{r}}{\'\i} and Derner, Erik and Babu{\v{s}}ka, Robert},
  journal={IEEE Access},
  year={2024},
  publisher={IEEE}
}

@article{meidani2023snip,
  title={SNIP: Bridging Mathematical Symbolic and Numeric Realms with Unified Pre-training},
  author={Meidani, Kazem and Shojaee, Parshin and Reddy, Chandan K and Farimani, Amir Barati},
  journal={arXiv preprint arXiv:2310.02227},
  year={2023}
}

@article{wu2023discovering,
  title={Discovering Mathematical Expressions Through {DeepSymNet}: A Classification-Based Symbolic Regression Framework},
  author={Wu, Min and Li, Weijun and Yu, Lina and Sun, Linjun and Liu, Jingyi and Li, Wenqiang},
  journal={IEEE Transactions on Neural Networks and Learning Systems},
  year={2023},
  publisher={IEEE}
}

@inproceedings{li2023transformer,
  title     = {Transformer-Based Model for Symbolic Regression via Joint Supervised Learning},
  author    = {Li, Wenqiang and Li, Weijun and Sun, Linjun and Wu, Min and Yu, Lina and Liu, Jingyi and Li, Yanjie and Tian, Songsong},
  booktitle = {The Eleventh International Conference on Learning Representations},
  year      = {2023},
  url       = {https://openreview.net/forum?id=ULzyv9M1j5}
}

@article{udrescu2020aifeynman,
  title   = {{AI Feynman}: A Physics-Inspired Method for Symbolic Regression},
  author  = {Udrescu, Silviu-Marian and Tegmark, Max},
  journal = {Science Advances},
  volume  = {6},
  number  = {16},
  pages   = {eaay2631},
  year    = {2020},
  doi     = {10.1126/sciadv.aay2631}
}

@inproceedings{udrescu2020aifeynman2,
  title     = {{AI Feynman} 2.0: Pareto-Optimal Symbolic Regression Exploiting Graph Modularity},
  author    = {Udrescu, Silviu-Marian and Tan, Andrew and Feng, Jiahai and Neto, Orisvaldo and Wu, Tailin and Tegmark, Max},
  booktitle = {Advances in Neural Information Processing Systems},
  volume    = {33},
  pages     = {4860--4871},
  year      = {2020}
}

@article{landajuela2022unified,
  title={A unified framework for deep symbolic regression},
  author={Landajuela, Mikel and Lee, Chak Shing and Yang, Jiachen and Glatt, Ruben and Santiago, Claudio P and Aravena, Ignacio and Mundhenk, Terrell and Mulcahy, Garrett and Petersen, Brenden K},
  journal={Advances in Neural Information Processing Systems},
  volume={35},
  pages={33985--33998},
  year={2022}
}

@inproceedings{holt2023deep,
  title     = {Deep Generative Symbolic Regression},
  author    = {Holt, Samuel and Qian, Zhaozhi and van der Schaar, Mihaela},
  booktitle = {The Eleventh International Conference on Learning Representations},
  year      = {2023},
  url       = {https://openreview.net/forum?id=o7koEEMA1bR}
}

@article{liu2023snr,
  title={{SNR}: Symbolic network-based rectifiable learning framework for symbolic regression},
  author={Liu, Jingyi and Li, Weijun and Yu, Lina and Wu, Min and Sun, Linjun and Li, Wenqiang and Li, Yanjie},
  journal={Neural Networks},
  volume={165},
  pages={1021--1034},
  year={2023},
  publisher={Elsevier}
}

@article{li2023metasymnet,
  title={{Metasymnet}: A dynamic symbolic regression network capable of evolving into arbitrary formulations},
  author={Li, Yanjie and Li, Weijun and Yu, Lina and Wu, Min and Liu, Jinyi and Li, Wenqiang and Hao, Meilan and Wei, Shu and Deng, Yusong},
  journal={arXiv preprint arXiv:2311.07326},
  year={2023}
}

@article{liu2025camo,
  title={{CaMo}: Capturing the modularity by end-to-end models for Symbolic Regression},
  author={Liu, Jingyi and Wu, Min and Yu, Lina and Li, Weijun and Li, Wenqiang and Li, Yanjie and Hao, Meilan and Deng, Yusong and Wei, Shu},
  journal={Knowledge-Based Systems},
  volume={309},
  pages={112747},
  year={2025},
  publisher={Elsevier}
}

@article{dong2025recent,
  title={Recent Advances in Symbolic Regression},
  author={Dong, Junlan and Zhong, Jinghui},
  journal={ACM Computing Surveys},
  volume={57},
  number={11},
  pages={1--37},
  year={2025},
  publisher={ACM New York, NY}
}

@article{al2024genetic,
  title={Genetic programming for feature selection based on feature removal impact in high-dimensional symbolic regression},
  author={Al-Helali, Baligh and Chen, Qi and Xue, Bing and Zhang, Mengjie},
  journal={IEEE Transactions on Emerging Topics in Computational Intelligence},
  volume={8},
  number={3},
  pages={2269--2282},
  year={2024},
  publisher={IEEE}
}

@article{shojaee2024llm,
  title={LLM-SR: Scientific Equation Discovery via Programming with Large Language Models},
  author={Shojaee, Parshin and Meidani, Kazem and Gupta, Shashank and Farimani, Amir Barati and Reddy, Chandan K},
  journal={arXiv preprint arXiv:2404.18400},
  year={2024}
}

@inproceedings{xu2024rsrm,
  title     = {Reinforcement Symbolic Regression Machine},
  author    = {Xu, Yilong and Liu, Yang and Sun, Hao},
  booktitle = {The Twelfth International Conference on Learning Representations},
  year      = {2024},
  url       = {https://openreview.net/forum?id=PJVUWpPnZC}
}

@article{li2025gptmcts,
  title   = {Discovering Mathematical Formulas from Data via {GPT}-Guided {Monte Carlo Tree Search}},
  author  = {Li, Yanjie and Li, Weijun and Yu, Lina and Wu, Min and Liu, Jingyi and Li, Wenqiang and Hao, Meilan},
  journal = {Expert Systems with Applications},
  volume  = {281},
  pages   = {127591},
  year    = {2025},
  doi     = {10.1016/j.eswa.2025.127591}
}

@misc{cranmer2023pysr,
  title         = {Interpretable Machine Learning for Science with {PySR} and {SymbolicRegression.jl}},
  author        = {Cranmer, Miles},
  year          = {2023},
  publisher     = {arXiv},
  eprint        = {2305.01582},
  archivePrefix = {arXiv},
  doi           = {10.48550/arXiv.2305.01582},
  url           = {https://arxiv.org/abs/2305.01582}
}

@article{virgolin2021improving,
  title={Improving model-based genetic programming for symbolic regression of small expressions},
  author={Virgolin, Marco and Alderliesten, Tanja and Witteveen, Cees and Bosman, Peter AN},
  journal={Evolutionary computation},
  volume={29},
  number={2},
  pages={211--237},
  year={2021},
  publisher={MIT Press One Rogers Street, Cambridge, MA 02142-1209, USA journals-info~…}
}

@article{xu2026nestynet,
  title={NestyNet. III. Symbolic Regression from Analytic Neural Surrogates},
  author={Xu, Yilong and Liu, Yang and Sun, Hao},
  journal={arXiv preprint arXiv:2608.21051},
  year={2026}
}
\bibliographystyle{iclr2027_conference}

\appendix

\section{Detailed definitions of the AI Feynman}

\subsection{Translational Symmetry}
For an $n$-dimensional function $f(X)$, where $X=(x_1,...,x_n)$ denotes the $n$-dimensional variables, if $f$ satisfies $f(x_i, x_j, ...) = f(x_i + x_j, ...)$ for any given $x_i, x_j$, then $f(X)$ is said to possess additive translational symmetry with respect to dimensions $x_i$ and $x_j$. Note that the left-hand side of the equation, $f(x_i, x_j, ...)$, involves (n) variables, whereas the right-hand side, $f(x_i+x_j, ...)$, involves only (n-1) variables. This means that $x_i$ and $x_j$ can be combined into a single new variable due to translational symmetry. Similar statements appearing later should be interpreted in the same manner. In the AI Feynman framework, for any $f(x_i, x_j, ...)$, if the condition $f(x_i, x_j, ...) = f(x_i + c, x_j - c, ...)$ holds for an arbitrary constant $c$, the function $f(X)$ is considered to exhibit additive translational symmetry. Analogous definitions apply to subtraction, multiplication, and division.

However, the AI Feynman method fails to effectively identify these symmetries in the presence of interference of constant coefficients or exponents, such as in the cases of $f(x_1 + 2 x_2)$ or $f(x_1 x_2^2)$.

\subsection{Variable Separability}
Consider an $n$-dimensional function $f(SX_1, SX_2)$, where $SX$ denotes a set consisting of an arbitrary number of variables $x_i$ ($i \in \{1,...,n\}$), and all such disjoint sets $SX$ span the $n$-dimensional space. If $f$ satisfies $f(SX_1, SX_2) = g(SX_1) + h(SX_2)$ for any $X$, then $f(X)$ is termed additively separable with respect to $SX_1$ and $SX_2$.  In AI Feynman, for any $f(SX_1, SX_2)$ and a constant $c$, if the relation $f(SX_1, SX_2) = f(SX_1, c) + f(c, SX_2) - f(c, c)$ is satisfied, $f(X)$ is deemed decomposable into $f(SX_1, SX_2) = g(SX_1) + h(SX_2)$. Multiplicative variable separability is determined via a similar criterion.

Nevertheless, this approach is incapable of effectively identifying separability in scenarios involving overlapping variables or interference of additive constant offset, such as $f(X) = g(x_1, x_2) + h(x_2, x_3)$ or $f(X) = g(x_1) h(x_2) + c$.

\subsection{Nested Composition}
\textbf{Compositionality}: An $n$-dimensional function $f(X)$ is said to possess compositionality if it can be formulated as $f(X) = g(h(X))$. AI Feynman postulates that if the gradient $\nabla f(X)$ aligns with the direction of $\nabla h(X)$ for all $X$, then $f(X)$ can be represented as $f(X) = g(h(X))$.

\textbf{Generalized Symmetry}: An $n$-dimensional function $f(SX_1, SX_2)$ exhibits generalized symmetry if it takes the form $f(X) = g(h(SX_1), SX_2)$. In AI Feynman, if the direction of the partial gradient $\nabla_{SX_1} f(SX_1, SX_2)$ is independent of $SX_2$ for all $X$, $f(X)$ is assumed to be representable as $f(X) = g(h(SX_1), SX_2)$.

\textbf{Generalized Additivity}: A 2-dimensional function $f(x_1, x_2)$ is characterized by generalized additivity if it can be expressed as $f(X) = F(g(x_1) + h(x_2))$. AI Feynman determines this property by checking whether the ratio of partial derivatives $\frac{\partial f / \partial x_1}{\partial f / \partial x_2}$ is multiplicatively separable with respect to $x_1$ and $x_2$.

The condition for compositionality requires evaluating $\nabla h(X)$, despite the analytical form of $h(X)$ being intrinsically unknown. The criterion for generalized additivity is not exclusively sensitive to additive structures and is prone to false positives when encountering multiplicative structures (a detailed proof is provided in \ref{Generalized Separable Forms Method} of Methodology and \ref{Generalized Separable Forms} of Appendix). Even if generalized symmetry or generalized additivity is successfully identified, these conditions do not yield the explicit forms of the inner functions $g$ and $h$. 

\section{Numerical Decision Criteria}
\label{Numerical Decision Criteria}

During the experiments, for the decision criteria derived from mathematical principles that theoretically require exact equality to zero, we set the threshold to $0.01$. Specifically, we define the relative loss as
\begin{equation}
    \mathrm{loss}_{\mathrm{relative}}
    =
    \frac{
        \left|\mathrm{loss}_{\mathrm{math}}\right|
    }{
        \max\limits_{x \in X_{\mathrm{all}}}|x|+\epsilon
    },
\end{equation}
where $\mathrm{loss}_{\mathrm{math}}$ denotes the numerical residual of the corresponding expression that is theoretically required to be equal to zero, and $X_{\mathrm{all}}$ denotes the set of all computational terms involved in the evaluation of $\mathrm{loss}_{\mathrm{math}}$. $\epsilon = 10^{-10}$ is introduced to avoid numerical instability when the denominator approaches zero. When $\mathrm{loss}_{\mathrm{relative}}$ is smaller than the predefined threshold, the corresponding criterion is considered satisfied.

For an unknown expression, we sequentially examine it according to the order of the proposed principles. If none of the principles is satisfied, or if valid values cannot be obtained during numerical sampling verification, the corresponding branch is terminated and returned. If a decomposition principle is successfully identified, the next-step computation is performed according to the formula of that principle, and the same decomposition procedure is recursively applied to the resulting sub-expressions until all branches can no longer be decomposed. This constitutes the expression decomposition process.

In the decomposition-based regression experiments, each terminal sub-expression at the end of every decomposition branch is processed by a symbolic regression algorithm once. The symbolic regression results of multiple sub-expressions are then recombined in reverse according to the decomposition structure to reconstruct the original complete expression.

It is worth noting that, for cases where NaN or Inf values occur during numerical evaluation, we perform finite resampling attempts. If valid sampled values cannot be obtained after these attempts, the corresponding criterion is considered invalid.

\section{Limitation and Future Work}
\label{Limitation and Future Work}
Methods of this kind, such as DDRSR and AI Feynman, are inherently grounded in rigorous mathematical principles. The implementation of these principles typically involves strict numerical criteria and therefore requires access to sufficiently accurate data, which constitutes an inherent limitation of such methods.

In practical applications, we usually have access only to an initial set of randomly sampled data points, whereas rigorous mathematical criteria often require information at specific critical coordinates. This discrepancy restricts the implementation and applicability of such algorithms. AI Feynman alleviates this issue by employing a surrogate model, namely a neural network, to approximate the underlying ground-truth mapping. However, this approach requires a large number of initial samples to adequately capture the fine-grained characteristics of the target domain. Alternatively, when an active sampling environment is available, such as a directly queryable physical field, accurate numerical values and gradient information can be efficiently obtained by combining direct sampling with numerical differentiation.

Moreover, in the presence of noise, sampling errors, or other perturbations, all sampled data may contain deviations. Consequently, quantities that are theoretically required to be exactly zero may take random nonzero values because of these deviations. This can interfere with the evaluation of the mathematical principles underlying methods such as DDRSR, thereby degrading their decomposition performance.

Nevertheless, when expression decomposition fails because of approximation errors, sampling biases, or other perturbations, the symbolic regression algorithm can still be applied directly to the undecomposed expression, which is equivalent to performing symbolic regression without expression decomposition. Conversely, once a candidate decomposition is successfully identified and verified, the expression can be simplified according to the corresponding decomposition structure. Therefore, decomposition algorithms based on rigorous mathematical principles are designed to avoid performance degradation when decomposition fails, while successful decomposition can provide additional performance gains.

These limitations also suggest two main directions for future development. First, additional mathematical principles with broader applicability, or principles tailored to specific application scenarios, may be incorporated to further extend the theoretical scope of the framework. Second, further improvements can focus on the practical implementation of these principles, including mitigating the effects of noise and sampling errors and developing more accurate surrogate models for reliable estimation of function values and derivative information.

\section{Mathematical Derivations and Proofs}
\label{Mathematical Derivations and Proofs}
\subsection{Additive Translational Symmetry}
\textbf{Derivation}: For an $n$-dimensional function $f(X)$, if $f(X)$ possesses additive translational symmetry with respect to dimensions $x_i$ and $x_j$, then $f(X)$ can be expressed as:
$$
f(x_i, x_j, SX) = f(c_1 \times x_i + c_2 \times x_j, SX)
$$
Here, $SX$ denotes the remaining dimensions excluding $x_i$ and $x_j$. By introducing a new variable $x_{ij} = c_1 \times x_i + c_2 \times x_j$, we obtain:
$$
\frac{\partial f(X)}{\partial x_i} = \frac{\partial f(X)}{\partial x_{ij}} \times \frac{\partial x_{ij}}{\partial x_i} = c_1 \times \frac{\partial f(X)}{\partial x_{ij}}$$$$\frac{\partial f(X)}{\partial x_j} = \frac{\partial f(X)}{\partial x_{ij}} \times \frac{\partial x_{ij}}{\partial x_j} = c_2 \times \frac{\partial f(X)}{\partial x_{ij}}
$$
Consequently, the ratio of these partial derivatives is given by:
$$
\frac{\partial f(X)}{\partial x_i} \div \frac{\partial f(X)}{\partial x_j} = \frac{c_1}{c_2} = \text{constant}
$$
If the above equation holds for any arbitrary $x_i$ and $x_j$, we determine that $f(X)$ possesses additive translational symmetry along the $x_i$ and $x_j$ dimensions, prompting the introduction of $x_{new} = x_i + \frac{c_2}{c_1} \times x_j$.  

\subsection{Multiplicative Translational Symmetry}
\label{Multiplicative Translational Symmetry}
\textbf{Derivation}: For an $n$-dimensional function $f(X)$, if $f(X)$ possesses multiplicative translational symmetry with respect to dimensions $x_i$ and $x_j$, then $f(X)$ can be formulated as:
$$
f(x_i, x_j, SX) = f(x_i^{c_1} \times x_j^{c_2}, SX)
$$
Here, $SX$ denotes the remaining dimensions excluding $x_i$ and $x_j$. By introducing a new variable $x_{ij} = x_i^{c_1} \times x_j^{c_2}$, we obtain:
$$
\frac{\partial f(X)}{\partial x_i} = \frac{\partial f(X)}{\partial x_{ij}} \times \frac{\partial x_{ij}}{\partial x_i} = c_1 \times \frac{\partial f(X)}{\partial x_{ij}} \times x_i^{c_1 - 1} \times x_j^{c_2}$$$$\frac{\partial f(X)}{\partial x_j} = \frac{\partial f(X)}{\partial x_{ij}} \times \frac{\partial x_{ij}}{\partial x_j} = c_2 \times \frac{\partial f(X)}{\partial x_{ij}} \times x_i^{c_1} \times x_j^{c_2 - 1}
$$
Consequently, dividing the two equations yields:
$$\frac{\partial f(X)}{\partial x_i} \div \frac{\partial f(X)}{\partial x_j} = \frac{c_1}{c_2} \times \frac{x_j}{x_i}$$$$\frac{\partial f(X)}{\partial x_i} \div \frac{\partial f(X)}{\partial x_j} \times \frac{x_i}{x_j} = \frac{c_1}{c_2} = \text{constant}
$$
If the above equation holds for any arbitrary $x_i$ and $x_j$, we determine that $f(X)$ possesses multiplicative translational symmetry along the $x_i$ and $x_j$ dimensions, prompting the introduction of $X_{new} = x_i^{c_1} \times x_j^{c_2}$.  

\subsection{Additive Separability with Overlapping Variables}
\label{Additive Separability with Overlapping Variables}
\textbf{Derivation}: For an $n$-dimensional function $f(X)$, if $f(X)$ exhibits additive separability with overlapping variables with respect to dimensions $SX_1$ and $SX_3$, then $f(X)$ can be expressed as:
$$
f(SX_1, SX_2, SX_3) = g(SX_1, SX_2) + h(SX_2, SX_3)
$$
By manipulating the terms, we derive:
$$
f(SX_1, SX_2, SX_3) = [g(SX_1, SX_2) + h(SX_2, SC_3)] + [g(SC_1, SX_2) + h(SX_2, SX_3)] - [g(SC_1, SX_2) + h(SX_2, SC_3)]
$$
$$= f(SX_1, SX_2, SC_3) + f(SC_1, SX_2, SX_3) - f(SC_1, SX_2, SC_3)
$$
If the above relationship holds for any $SX_1$ and $SX_3$ alongside any fixed constants $SC$, we conclude that $f(X)$ is additively separable across dimensions $SX_1$ and $SX_3$. To obtain the explicit function mappings for the sub-functions $g$ and $h$ to facilitate further separation, we define:
$$
g(SX_1, SX_2) = f(SX_1, SX_2, SC_3) - f(SC_1, SX_2, SC_3)$$$$h(SX_2, SX_3) = f(SC_1, SX_2, SX_3)
$$

\subsection{Multiplicative Separability with Overlapping Variables}
\label{Multiplicative Separability with Overlapping Variables}
\textbf{Derivation}: For an $n$-dimensional function $f(X)$, if $f(X)$ exhibits multiplicative separability with overlapping variables with respect to dimensions $SX_1$ and $SX_3$, then $f(X)$ can be expressed as:
$$
f(SX_1, SX_2, SX_3) = g(SX_1, SX_2) \times h(SX_2, SX_3)
$$
Through algebraic substitution, we establish:
$$
f(SX_1, SX_2, SX_3) = \frac{[g(SX_1, SX_2) \times h(SX_2, SC_3)] \times [g(SC_1, SX_2) \times h(SX_2, SX_3)]}{[g(SC_1, SX_2) \times h(SX_2, SC_3)]}$$$$= \frac{f(SX_1, SX_2, SC_3) \times f(SC_1, SX_2, SX_3)}{f(SC_1, SX_2, SC_3)}
$$
If the above relationship holds for any $SX_1$ and $SX_3$ alongside any fixed constants $SC$, we conclude that $f(X)$ is multiplicatively separable across dimensions $SX_1$ and $SX_3$. To obtain the function mappings for sub-functions $g$ and $h$ to enable further decomposition, we define:
$$
g(SX_1, SX_2) = \frac{f(SX_1, SX_2, SC_3)}{f(SC_1, SX_2, SC_3)}$$$$h(SX_2, SX_3) = f(SC_1, SX_2, SX_3)
$$

\subsection{Compositional Form of Additive Separability}
\label{Compositional Form of Additive Separability}
\textbf{Proof}: Assume the function $f$ is additively separable with respect to dimensions $SX_1$ and $SX_3$, and concurrently additively separable with respect to $SX_2$ and $SX_3$. The function $f$ can therefore be represented as:
$$
f(SX_1, SX_2, SX_3) = g_1(SX_1, SX_2) + h_1(SX_2, SX_3) = g_2(SX_1, SX_2) + h_2(SX_1, SX_3)
$$
Taking the partial derivative with respect to $SX_3$ yields:
$$
\frac{\partial f(SX_1, SX_2, SX_3)}{\partial SX_3} = \frac{\partial h_1(SX_2, SX_3)}{\partial SX_3} = \frac{\partial h_2(SX_1, SX_3)}{\partial SX_3}
$$
Let $T(X) = h_1(SX_2, SX_3) - h_2(SX_1, SX_3)$. It follows that:
$$
\frac{\partial T(X)}{\partial SX_3} = \frac{\partial h_1(SX_2, SX_3)}{\partial SX_3} - \frac{\partial h_2(SX_1, SX_3)}{\partial SX_3} = 0
$$
This indicates that $T(X)$ is independent of $SX_3$ and can be strictly expressed in the form $T(X) = T(SX_1, SX_2)$.  Simultaneously, we have:
$$\frac{\partial \frac{\partial T(SX_1, SX_2)}{\partial SX_1}}{\partial SX_2} = \frac{\partial \frac{\partial h_2(SX_1, SX_3)}{\partial SX_1}}{\partial SX_2} = 0, \quad \frac{\partial \frac{\partial T(SX_1, SX_2)}{\partial SX_2}}{\partial SX_1} = \frac{\partial \frac{\partial h_1(SX_2, SX_3)}{\partial SX_2}}{\partial SX_1} = 0
$$
Thus, $T(X)$ can be further decomposed into the form $T(X) = A(SX_1) + B(SX_2)$. Furthermore:
$$
h_1(SX_2, SX_3) - B(SX_2) = h_2(SX_1, SX_3) + A(SX_1)
$$
Given that the left side is a function exclusively of $SX_2$ and $SX_3$, and the right side is a function exclusively of $SX_1$ and $SX_3$, both sides must equal a function solely dependent on $SX_3$. We therefore deduce:
$$
h_1(SX_2, SX_3) = C(SX_3) + B(SX_2), \quad h_2(SX_1, SX_3) = C(SX_3) - A(SX_1)
$$
Substituting this back yields:
$$
f(SX_1, SX_2, SX_3) = g_1(SX_1, SX_2) + h_1(SX_2, SX_3) = [g_1(SX_1, SX_2) + B(SX_2)] + C(SX_3)
$$
This rigorously proves that $f$ admits a representation of the form $f(SX_1, SX_2, SX_3) = g(SX_1, SX_2) + h(SX_3)$.  

\subsection{Compositional Form of Multiplicative Separability}
\label{Compositional Form of Multiplicative Separability}
\textbf{Proof:}
Assume that the function $f$ is multiplicatively separable with respect to dimensions $SX_1$ and $SX_3$, and concurrently multiplicatively separable with respect to $SX_2$ and $SX_3$. The function $f$ can therefore be represented as:
\begin{equation}
\begin{aligned}
f(SX_1,SX_2,SX_3)
&= g_1(SX_1,SX_2)\times h_1(SX_2,SX_3) \\
&= g_2(SX_1,SX_2)\times h_2(SX_1,SX_3).
\end{aligned}
\end{equation}

The following derivation is conducted on the considered subdomain where the relevant factors and denominators are well defined. We fix $SX_1$ to an arbitrary valid constant $SC_1$. It follows that:
\begin{equation}
g_1(SC_1,SX_2)\times h_1(SX_2,SX_3)
=
g_2(SC_1,SX_2)\times h_2(SC_1,SX_3).
\end{equation}

Rearranging the above equation yields:
\begin{equation}
h_1(SX_2,SX_3)
=
\frac{g_2(SC_1,SX_2)}
{g_1(SC_1,SX_2)}
\times h_2(SC_1,SX_3).
\end{equation}

Let
\begin{equation}
q(SX_2)
=
\frac{g_2(SC_1,SX_2)}
{g_1(SC_1,SX_2)},
\end{equation}
and
\begin{equation}
h(SX_3)=h_2(SC_1,SX_3).
\end{equation}

We therefore obtain:
\begin{equation}
h_1(SX_2,SX_3)=q(SX_2)\times h(SX_3).
\end{equation}

Substituting this relationship back into the original expression gives:
\begin{equation}
\begin{aligned}
f(SX_1,SX_2,SX_3)
&=g_1(SX_1,SX_2)\times h_1(SX_2,SX_3) \\
&=g_1(SX_1,SX_2)\times q(SX_2)\times h(SX_3).
\end{aligned}
\end{equation}

Defining
\begin{equation}
g(SX_1,SX_2)
=
g_1(SX_1,SX_2)\times q(SX_2),
\end{equation}
we finally obtain:
\begin{equation}
f(SX_1,SX_2,SX_3)
=
g(SX_1,SX_2)\times h(SX_3).
\end{equation}

This rigorously proves that $f$ can be expressed in the form
\begin{equation}
f(SX_1,SX_2,SX_3)
=
g(SX_1,SX_2)\times h(SX_3).
\end{equation}

\subsection{Multiplicative Separability with Additive Constant Offset}
This criterion is applied after standard additive and multiplicative separability have been excluded. In addition, we require \(\frac{\partial^2 f(X)}{\partial x_i\partial x_j}\neq 0\) on the considered subdomain to avoid degenerate cases.

\label{Multiplicative Separability with Additive Constant Offset}
\textbf{Derivation}: For an $n$-dimensional function $f(X)$, if $f(X)$ exhibits multiplicative separability containing a constant term across dimensions $SX$, it can be expressed as:
$$
f(X) = g(x_i, SX) \times h(x_j, SX) + c
$$
Here, $SX$ denotes the remaining dimensions excluding $x_i$ and $x_j$. The partial derivatives are:
$$
\frac{\partial f(X)}{\partial x_i} = h(x_j, SX) \times \frac{\partial g(x_i, SX)}{\partial x_i}, \quad \frac{\partial f(X)}{\partial x_j} = g(x_i, SX) \times \frac{\partial h(x_j, SX)}{\partial x_j}$$$$\frac{\partial^2 f(X)}{\partial x_i \partial x_j} = \frac{\partial g(x_i, SX)}{\partial x_i} \times \frac{\partial h(x_j, SX)}{\partial x_j}
$$
By computing the ratios, we obtain:
$$\frac{\partial^2 f(X)}{\partial x_i \partial x_j} \div \frac{\partial f(X)}{\partial x_i} = \frac{\partial h(x_j, SX)}{\partial x_j} \div h(x_j, SX), \quad \frac{\partial^2 f(X)}{\partial x_i \partial x_j} \div \frac{\partial f(X)}{\partial x_j} = \frac{\partial g(x_i, SX)}{\partial x_i} \div g(x_i, SX)
$$
Let $u_i(X) = \frac{\partial^2 f(X)}{\partial x_i \partial x_j} \div \frac{\partial f(X)}{\partial x_i}$ and $u_j(X) = \frac{\partial^2 f(X)}{\partial x_i \partial x_j} \div \frac{\partial f(X)}{\partial x_j}$. If $f(X)$ exhibits multiplicative separability with a constant term, $u_i(X)$ is independent of $x_i$, and $u_j(X)$ is independent of $x_j$. Thus, $\frac{\partial u_i(X)}{\partial x_i} = 0$ and $\frac{\partial u_j(X)}{\partial x_j} = 0$.

Therefore, the above derivation establishes that multiplicative separability with a constant term necessarily satisfies the proposed differential criterion. In the practical decomposition procedure, this criterion is used as a candidate screening mechanism rather than a standalone proof of separability. When the criterion is satisfied, the corresponding constant term is further recovered and verified using the analytical procedure described in Appendix \ref{Separation Mechanism for Multiplicative Forms with Additive Constant Offset}. Only when the recovered constant term remains invariant under different values of the remaining variables is the structure accepted as a multiplicative separable form with a constant term.

\subsection{Separation Mechanism for Multiplicative Forms with Additive Constant Offset}
\label{Separation Mechanism for Multiplicative Forms with Additive Constant Offset}
When \(f(X)\) satisfies the proposed differential criterion for multiplicative separability with a constant term, we further recover and verify the constant term.

\textbf{Derivation}: Assuming that the candidate structure corresponds to $f(X) = g(x_i, SX) \times h(x_j, SX) + c(SX)$, where $c(SX)$ denotes a residual term that is independent of $x_i$ and $x_j$, but may depend on the remaining variables $S_X$. First, we fix all dimensions excluding $x_i$ and $x_j$ to valid random values, establishing a 2-dimensional subspace spanned by $x_i$ and $x_j$. We sample two valid points along the $x_i$ and $x_j$ dimensions, denoted as $(x_{i1}, x_{i2})$ and $(x_{j1}, x_{j2})$ respectively, forming a 2-dimensional bounding rectangle.  Within this subspace (where $SX$ acts as a constant, and is thus omitted from the subsequent notation for brevity), we evaluate the integrals:
$$
f(x_{i2}, x_{j2}) = f(x_{i1}, x_{j2}) + \int_{x_{i1}}^{x_{i2}} \frac{\partial f(X)}{\partial x_i}\bigg|_{x_j=x_{j2}} = f(x_{i1}, x_{j2}) + h(x_{j2}) \times \int_{x_{i1}}^{x_{i2}} \frac{\partial g(x_i)}{\partial x_i}$$$$f(x_{i2}, x_{j1}) = f(x_{i1}, x_{j1}) + h(x_{j1}) \times \int_{x_{i1}}^{x_{i2}} \frac{\partial g(x_i)}{\partial x_i}
$$
Therefore, by computing the ratio, we derive:
$$
\frac{h(x_{j2})}{h(x_{j1})} = \frac{f(x_{i2}, x_{j2}) - f(x_{i1}, x_{j2})}{f(x_{i2}, x_{j1}) - f(x_{i1}, x_{j1})}
$$
By identical logic, we deduce:
$$\frac{g(x_{i2})}{g(x_{i1})} = \frac{f(x_{i2}, x_{j2}) - f(x_{i2}, x_{j1})}{f(x_{i1}, x_{j2}) - f(x_{i1}, x_{j1})}
$$
Furthermore, we establish the difference:
$$
f(x_{i2}, x_{j2}) - f(x_{i2}, x_{j1}) = (g(x_{i2}) \times h(x_{j2}) + c) - (g(x_{i2}) \times h(x_{j1}) + c)
$$
Which yields:
$$
h(x_{j2}) - h(x_{j1}) = \frac{f(x_{i2}, x_{j2}) - f(x_{i2}, x_{j1})}{g(x_{i2})}
$$
Similarly:
$$
g(x_{i2}) - g(x_{i1}) = \frac{f(x_{i2}, x_{j2}) - f(x_{i1}, x_{j2})}{h(x_{j2})}
$$
By solving this system of simultaneous equations, we can analytically extract:
$$
g(x_{i2}) \times h(x_{j2}) = \frac{(f(x_{i2}, x_{j2}) - f(x_{i2}, x_{j1})) \times (f(x_{i2}, x_{j2}) - f(x_{i1}, x_{j2}))}{f(x_{i2}, x_{j2}) - f(x_{i1}, x_{j2}) - f(x_{i2}, x_{j1}) + f(x_{i1}, x_{j1})}
$$
Substituting this back into the original expression for $f(x_{i2}, x_{j2})$, the precise value of the constant $c$ is computed as:
$$
c(SX) = f(x_{i2}, x_{j2}) - \frac{(f(x_{i2}, x_{j2}) - f(x_{i2}, x_{j1})) \times (f(x_{i2}, x_{j2}) - f(x_{i1}, x_{j2}))}{f(x_{i2}, x_{j2}) - f(x_{i1}, x_{j2}) - f(x_{i2}, x_{j1}) + f(x_{i1}, x_{j1})}
$$

It should be emphasized that the above derivation is performed within a two-dimensional subspace with fixed \(SX\), and therefore the recovered residual is initially denoted as $c(SX)$. We repeat the recovery procedure for different values of $SX$. If

$$
c(SX^{(1)})=c(SX^{(2)})=\cdots=c, 
$$

the residual is independent of all remaining variables and is therefore a global constant. Only in this case is the criterion for multiplicative separability with a constant term accepted. Once this invariance is verified, we denote the common value of $c(SX)$ by the global constant $c$.

Finally, by setting $f_{new}(X) = f(X) - c$, the problem is mathematically reduced to standard multiplicative separability, thereby enabling the execution of subsequent variable separation procedures.  

\subsection{Generalized Separable Forms}
\label{Generalized Separable Forms}
\textbf{Derivation}: For an $n$-dimensional function $f(X)$, if $f(X)$ exhibits a generalized separable form with respect to dimensions $SX_1$ and $SX_3$, it can be expressed as:
$$
f(X) = U(g(x_i, SX) + h(x_j, SX)) \quad \text{or} \quad f(X) = U(g(x_i, SX) \times h(x_j, SX))
$$
Here, $SX$ denotes the remaining dimensions excluding $x_i$ and $x_j$. If $f(X)$ assumes the additive internal form $f(X) = U(g(x_i, SX) + h(x_j, SX))$, applying the chain rule yields:
$$
\frac{\partial f(X)}{\partial x_i} = \frac{\partial f(X)}{\partial (g(x_i, SX) + h(x_j, SX))} \times \frac{\partial g(x_i, SX)}{\partial x_i}
$$
$$
\frac{\partial f(X)}{\partial x_j} = \frac{\partial f(X)}{\partial (g(x_i, SX) + h(x_j, SX))} \times \frac{\partial h(x_j, SX)}{\partial x_j}
$$
$$
\frac{\partial f(X)}{\partial x_i} \div \frac{\partial f(X)}{\partial x_j} = \frac{\partial g(x_i, SX)}{\partial x_i} \div \frac{\partial h(x_j, SX)}{\partial x_j}
$$
Let $v(X) = \frac{\partial f(X)}{\partial x_i} \div \frac{\partial f(X)}{\partial x_j}$. It is evident that $v(X)$ is multiplicatively separable across dimensions $x_i$ and $x_j$.  Similarly, if $f(X)$ assumes the multiplicative internal form $f(X) = U(g(x_i, SX) \times h(x_j, SX))$, we have:  $$v(X) = \frac{\partial f(X)}{\partial x_i} \div \frac{\partial f(X)}{\partial x_j} = \frac{\partial g(x_i, SX)}{\partial x_i} \times h(x_j, SX) \div \left(\frac{\partial h(x_j, SX)}{\partial x_j} \times g(x_i, SX)\right)$$In this scenario, $v(X)$ is equally multiplicatively separable across dimensions $x_i$ and $x_j$.

Therefore, for any unknown function $f(X)$, we construct the metric $v(X)$ according to the definitions above. If $v(X)$ satisfies multiplicative separability across dimensions $x_i$ and $x_j$ for any $X$, we identify $f(X)$ as possessing a generalized separable form.

\subsection{Nested Separation of Standard Logarithmic and Exponential Functions}
\label{Nested Separation of Standard Logarithmic and Exponential Functions}
\textbf{Derivation}: For an $n$-dimensional function $f(X)$, if $f(X)$ possesses a nested separation form of a standard logarithmic function, it can be expressed as $f(X) = \log(g(X))$, where $g(X)$ can be subjected to further separation. 

Let $f_{new}(X) = \exp(f(X)) = g(X)$. Under this mapping, $f_{new}$ inherently maintains the capacity for further separation. Therefore, for any unknown target mapping $f(X)$, if $f_{new}(X)$ is proven to be separable, we confirm that the original function $f(X)$ possesses a nested separation form of a standard logarithmic function.  The derivation and logic for exponential functions ($\exp$) follow an identical paradigm.

It is worth noting that when evaluating the internal separability during this nested decomposition, we typically prohibit subsequent nested separation to prevent infinite recursive calls. Theoretically, however, guided by specific priors, the maximum nesting depth can be established as a tunable hyperparameter, thereby permitting a predefined number of nested hierarchical layers.

\subsection{Nested Separation Form of Standard Trigonometric Functions}
\label{Nested Separation Form of Standard Trigonometric Functions}
\textbf{Derivation}: For an $n$-dimensional function $f(X)$, if $f(X)$ possesses a nested separation form of standard trigonometric functions, it can be expressed as:
$$
f(X) = \sin(\text{or } \cos)(g(x_i, SX) + h(x_j, SX) + c) \quad \text{or} \quad f(X) = \sin(\text{or } \cos)(g(x_i, SX) \times h(x_j, SX) + c)
$$
Here, $SX$ denotes the remaining dimensions excluding $x_i$ and $x_j$. If $f(X)$ assumes the internal additive structure $f(X) = \sin(g(x_i, SX) + h(x_j, SX) + c)$, computing the partial derivatives yields:
$$\frac{\partial f(X)}{\partial x_i} = \frac{\partial f(X)}{\partial g(x_i, SX)} \times \frac{\partial g(x_i, SX)}{\partial x_i} = \cos(g(x_i, SX) + h(x_j, SX) + c) \times \frac{\partial g(x_i, SX)}{\partial x_i}$$$$\frac{\partial f(X)}{\partial x_j} = \cos(g(x_i, SX) + h(x_j, SX) + c) \times \frac{\partial h(x_j, SX)}{\partial x_j}
$$
Further multiplying these terms produces:
$$\frac{\partial f(X)}{\partial x_i} \times \frac{\partial f(X)}{\partial x_j} = \cos^2(g(x_i, SX) + h(x_j, SX) + c) \times \frac{\partial g(x_i, SX)}{\partial x_i} \times \frac{\partial h(x_j, SX)}{\partial x_j}
$$
$$
= (1 - f^2(X)) \times \frac{\partial g(x_i, SX)}{\partial x_i} \times \frac{\partial h(x_j, SX)}{\partial x_j}
$$
Let $T(X) = \frac{\partial f(X)}{\partial x_i} \times \frac{\partial f(X)}{\partial x_j} \div (1 - f^2(X))$. If $f(X)$ conforms to the specified structure, $T(X) = \frac{\partial g(x_i, SX)}{\partial x_i} \times \frac{\partial h(x_j, SX)}{\partial x_j}$ is demonstrably multiplicatively separable with respect to dimensions $x_i$ and $x_j$.  Analogously, assuming the function $f$ takes the internal multiplicative form $f(X) = \sin(g(x_i, SX) \times h(x_j, SX) + c)$, we derive:
$$
\frac{\partial f(X)}{\partial x_i} = \frac{\partial f(X)}{\partial g(x_i, SX)} \times \frac{\partial g(x_i, SX)}{\partial x_i} = \cos(g(x_i, SX) \times h(x_j, SX) + c) \times h(x_j, SX) \times \frac{\partial g(x_i, SX)}{\partial x_i}$$$$\frac{\partial f(X)}{\partial x_j} = \cos(g(x_i, SX) \times h(x_j, SX) + c) \times g(x_i, SX) \times \frac{\partial h(x_j, SX)}{\partial x_j}
$$
Further multiplying these terms produces:
$$
\frac{\partial f(X)}{\partial x_i} \times \frac{\partial f(X)}{\partial x_j} = \cos^2(g(x_i, SX) \times h(x_j, SX) + c) \times \left[g(x_i, SX) \times \frac{\partial g(x_i, SX)}{\partial x_i}\right] \times \left[h(x_j, SX) \times \frac{\partial h(x_j, SX)}{\partial x_j}\right]
$$
$$
= (1 - f^2(X)) \times \left[g(x_i, SX) \times \frac{\partial g(x_i, SX)}{\partial x_i}\right] \times \left[h(x_j, SX) \times \frac{\partial h(x_j, SX)}{\partial x_j}\right]
$$
Let $T(X) = \frac{\partial f(X)}{\partial x_i} \times \frac{\partial f(X)}{\partial x_j} \div (1 - f^2(X))$. If $f(X)$ conforms to the specified structure, $T(X) = \left[g(x_i, SX) \times \frac{\partial g(x_i, SX)}{\partial x_i}\right] \times \left[h(x_j, SX) \times \frac{\partial h(x_j, SX)}{\partial x_j}\right]$ is multiplicatively separable across dimensions $x_i$ and $x_j$.

The derivation for the cosine function ($\cos$) follows an identical mathematical progression. Therefore, for an unknown function $f(X)$, we construct the metric $T(X)$. When $T(X)$ is verified to be multiplicatively separable across any two arbitrary dimensions, the function $f(X)$ is considered to exhibit a standard trigonometric nested separation form.

It is critical to note that, due to the inherent periodicity of trigonometric operators and the uncertainty regarding the internal sub-functions, acquiring explicit deterministic mappings for the sub-functions is analytically intractable. Consequently, upon identifying this special nested form for $\sin$ (or $\cos$), the algorithm does not proceed with further variable separation along these nodes.

\subsection{Non-identifiability of Decomposition Without Variable Separability}
\label{Non-identifiability of Decomposition Without Variable Separability}
\textbf{Proof}: Assume a given function $f$ structurally satisfies the following form:
$$
f(SX_1, SX_2) = g(SX_1, SX_2) \textcircled{+} h(SX_2)
$$
where $\textcircled{+}$ denotes an associative and commutative binary operation with an identity element and an inverse operation $\textcircled{-}$. Then for any arbitrary function $t(SX_2)$ defined over the subspace $SX_2$, the following identity holds:
$$
f(SX_1, SX_2) = g(SX_1, SX_2) \textcircled{+} h(SX_2) \textcircled{+} t(SX_2) \textcircled{-} t(SX_2)
$$
$$
= (g(SX_1, SX_2) \textcircled{-} t(SX_2)) \textcircled{+} (h(SX_2) \textcircled{+} t(SX_2)) = g_t(SX_1, SX_2) \textcircled{+} h_t(SX_2)
$$
This rigorously demonstrates that an infinite number of valid functional configurations exist for $g(SX_1, SX_2)$ and $h(SX_2)$, rendering identifiable decomposition fundamentally impossible under these unconstrained conditions.

\section{Hyperparameter configurations of symbolic regression methods.}
\begin{longtable}{p{3.5cm} p{6cm} p{5.5cm}}
    \caption{Hyperparameter configurations of symbolic regression methods.}
    \label{tab:hyperparameter} \\
    \toprule
    \textbf{Method} & \textbf{Hyperparameter} & \textbf{Value} \\
    \midrule
    \endfirsthead
    
    \multicolumn{3}{c}%
    {{\bfseries \tablename\ \thetable{} -- continued from previous page}} \\
    \toprule
    \textbf{Method} & \textbf{Hyperparameter} & \textbf{Value} \\
    \midrule
    \endhead

    \midrule \multicolumn{3}{r}{{Continued on next page}} \\
    \endfoot

    \bottomrule
    \endlastfoot

    \textbf{Hardware}
    & GPU & four NVIDIA GV100(TITAN V, 12GB memory) \\
    & CPU & Intel(R) Xeon(R) Gold 6138 CPU @ 2.00GHz(40 cores) \\
    \midrule
    
    \textbf{General} 
    & DDR sample variable number & 20 \\
    & DDR sample constant number & 10 \\
    & SR sample number & 40 \\
    & SR test number & 40 \\
    & variable sample range & (0,1) \\
    & difference step size & 1e-5 \\
\end{longtable}
\section{Expressions of Datasets}
\footnotesize
\setlength{\tabcolsep}{6pt}
\begin{longtable}{l c p{10cm}}
\caption{Expressions of Datasets} \label{tab:Dataset_Expressions} \\
\toprule
\textbf{Dataset} & \textbf{Variables} & \textbf{Formula} \\
\midrule
\endfirsthead
\multicolumn{3}{c}{{\tablename\ \thetable{} -- continued from previous page}} \\
\toprule
\textbf{Dataset} & \textbf{Variables} & \textbf{Formula} \\
\midrule
\endhead
\midrule
\multicolumn{3}{r}{{Continued on next page}} \\
\endfoot
\bottomrule
\endlastfoot
Keijzer-10 & 2 & \texttt{pow(x1,\allowbreak x2)\allowbreak } \\
Keijzer-11 & 2 & \texttt{x1*\allowbreak x2+\allowbreak sin((x1-\allowbreak 1)\allowbreak *\allowbreak (x2-\allowbreak 1)\allowbreak )\allowbreak } \\
Keijzer-12 & 2 & \texttt{pow(x1,\allowbreak 4)\allowbreak -\allowbreak pow(x1,\allowbreak 3)\allowbreak +\allowbreak div(pow(x2,\allowbreak 2)\allowbreak ,\allowbreak 2)\allowbreak -\allowbreak x2} \\
Keijzer-13 & 2 & \texttt{6*\allowbreak sin(x1)\allowbreak *\allowbreak cos(x2)\allowbreak } \\
Keijzer-14 & 2 & \texttt{div(8,\allowbreak 2+\allowbreak pow(x1,\allowbreak 2)\allowbreak +\allowbreak pow(x2,\allowbreak 2)\allowbreak )\allowbreak } \\
Keijzer-15 & 2 & \texttt{div(pow(x1,\allowbreak 3)\allowbreak ,\allowbreak 5)\allowbreak +\allowbreak div(pow(x2,\allowbreak 3)\allowbreak ,\allowbreak 2)\allowbreak -\allowbreak x2-\allowbreak x1} \\
Nguyen-9 & 2 & \texttt{sin(x1)\allowbreak +\allowbreak sin(pow(x2,\allowbreak 2)\allowbreak )\allowbreak } \\
Nguyen-10 & 2 & \texttt{2*\allowbreak sin(x1)\allowbreak *\allowbreak cos(x2)\allowbreak } \\
\end{longtable}
\section{Statistical Experimental Results}
\begin{table*}[ht]
\centering
\caption{Separation Score of DDRSR and AI Feynman in comprehensive experiments.}
\label{tab:Comprehensive_score}
\begin{tabular}{lcc}
\toprule
Dataset & DDRSR & AI Feynman \\
\midrule
Keijzer & 0.6666 & 0.5000 \\
Nguyen & 1.0000 & 0.8000 \\
Constant & 1.0000 & 0.6666 \\
Livermore & 0.8333 & 0.6666 \\
Jin & 0.8333 & 0.8333 \\
strogatz & 0.4285 & 0.3571 \\
Korns & 1.0000 & 0.3888 \\
Neat & 1.0000 & 0.7500 \\
Vladislavleva & 0.8571 & 0.7142 \\
Others & 0.7500 & 0.5000 \\
Feynman & 0.9685 & 0.7842 \\
Simple\_Test & 1.0000 & 0.5909 \\
Complex-Test & 1.0000 & 0.2222 \\
\bottomrule
\end{tabular}
\end{table*}
\begin{table*}[ht]
\centering
\caption{Performance ($R^2$ and Recovery Rate) across different number of variables in comprehensive experiments.}
\label{tab:Comprehensive_variable_R2Rec}
\begin{small}
\begin{tabular}{lll ccccc}
\toprule
\multirow{2}{*}{Method} & \multirow{2}{*}{Metric} & \multirow{2}{*}{Variant} & \multicolumn{5}{c}{Number of Variables} \\
\cmidrule(lr){4-8}
& & & 2 & 3 & 4 & 5 & 6+ \\
\midrule
\multirow{6}{*}{GP-GOMEA} & \multirow{3}{*}{$R^2$} & Orig. & 0.9664 & 0.9441 & 0.8315 & 0.8190 & -0.1853 \\
 & & AIF. & 0.9918 & 0.9410 & 0.9252 & 0.9853 & 0.5313 \\
 & & DDRSR & 0.9957 & 0.9563 & 0.9798 & 0.9946 & 0.8526 \\
\cmidrule(lr){2-8}
 & \multirow{3}{*}{Rec.} & Orig. & 0.3575 & 0.4772 & 0.2514 & 0.1400 & 0.0285 \\
 & & AIF. & 0.7393 & 0.8136 & 0.7714 & 0.6300 & 0.4285 \\
 & & DDRSR & 0.7969 & 0.8636 & 0.8285 & 0.8300 & 0.6857 \\
\midrule
\multirow{6}{*}{E2E} & \multirow{3}{*}{$R^2$} & Orig. & 0.9386 & 0.8129 & 0.6418 & 0.3675 & 0.0405 \\
 & & AIF. & 0.9758 & 0.9449 & 0.7755 & 0.8188 & 0.6134 \\
 & & DDRSR & 0.9912 & 0.9659 & 0.9241 & 0.9518 & 0.8529 \\
\cmidrule(lr){2-8}
 & \multirow{3}{*}{Rec.} & Orig. & 0.0666 & 0.0181 & 0.0000 & 0.0000 & 0.0000 \\
 & & AIF. & 0.1666 & 0.2045 & 0.1257 & 0.1200 & 0.0142 \\
 & & DDRSR & 0.1909 & 0.3772 & 0.2457 & 0.2000 & 0.3428 \\
\midrule
\multirow{6}{*}{PySR} & \multirow{3}{*}{$R^2$} & Orig. & 0.9974 & 0.9776 & 0.8466 & 0.7466 & -0.2700 \\
 & & AIF. & 0.9933 & 0.9882 & 0.9246 & 0.9175 & 0.7835 \\
 & & DDRSR & 0.9997 & 0.9934 & 0.9779 & 0.9833 & 0.9843 \\
\cmidrule(lr){2-8}
 & \multirow{3}{*}{Rec.} & Orig. & 0.6818 & 0.7500 & 0.5428 & 0.5000 & 0.2142 \\
 & & AIF. & 0.7272 & 0.8181 & 0.6000 & 0.6500 & 0.5714 \\
 & & DDRSR & 0.8030 & 0.8863 & 0.6857 & 0.7500 & 0.7142 \\
\bottomrule
\end{tabular}
\end{small}
\end{table*}

\clearpage
\begin{table*}[t]
\centering
\caption{Performance ($R^2$ and Recovery Rate) across different datasets in comprehensive experiments.}
\label{tab:Comprehensive_dataset_R2Rec}
\begin{small}
\setlength{\tabcolsep}{3pt}
\begin{tabular}{@{} lllccc @{\hspace{2em}} lllccc @{}}
\toprule
Dataset & Metric & Variant & E2E & GP-GOMEA & PySR & Dataset & Metric & Variant & E2E & GP-GOMEA & PySR \\
\cmidrule(lr){1-6} \cmidrule(lr){7-12}

\multirow{6}{*}{Keijzer} & \multirow{3}{*}{$R^2$} & Orig. & 0.9382 & 0.9896 & 0.9946 & \multirow{6}{*}{Nguyen} & \multirow{3}{*}{$R^2$} & Orig. & 0.9507 & 0.9936 & 0.9960 \\
 &  & AIF. & 0.9830 & 0.9940 & 0.9971 &  &  & AIF. & 0.9864 & 0.9999 & 0.9981 \\
 &  & DDRSR & 0.9876 & 0.9949 & 0.9995 &  &  & DDRSR & 0.9993 & 1.0000 & 0.9996 \\
\cmidrule(lr){2-6} \cmidrule(lr){8-12}
 & \multirow{3}{*}{Rec.} & Orig. & 0.0000 & 0.3000 & 0.6666 &  & \multirow{3}{*}{Rec.} & Orig. & 0.0000 & 0.2000 & 0.8000 \\
 &  & AIF. & 0.0000 & 0.6666 & 0.1000 &  &  & AIF. & 0.0000 & 0.9600 & 0.5200 \\
 &  & DDRSR & 0.0000 & 0.6666 & 0.2333 &  &  & DDRSR & 0.1200 & 1.0000 & 0.5600 \\
\cmidrule(lr){1-6} \cmidrule(lr){7-12}

\addlinespace[0.5em]
\multirow{6}{*}{Constant} & \multirow{3}{*}{$R^2$} & Orig. & 0.9633 & 0.9978 & 0.9212 & \multirow{6}{*}{Livermore} & \multirow{3}{*}{$R^2$} & Orig. & 0.9743 & 0.9673 & 0.9973 \\
\bottomrule
\end{tabular}
\end{small}
\end{table*}

\clearpage
\begin{table*}[ht]
\centering
\caption{Separation Score of DDRSR and AI Feynman in Ablation Experiments of Expression Separation.}
\label{tab:Separation_score}
\begin{tabular}{lcc}
\toprule
Dataset & DDRSR & AI Feynman \\
\midrule
Keijzer & 0.6666 & 0.5000 \\
Nguyen & 1.0000 & 0.8000 \\
Constant & 1.0000 & 0.6666 \\
Livermore & 0.8333 & 0.6666 \\
Jin & 0.8333 & 0.8333 \\
strogatz & 0.4285 & 0.3571 \\
Korns & 1.0000 & 0.1111 \\
Neat & 1.0000 & 0.7500 \\
Vladislavleva & 0.8571 & 0.7142 \\
Others & 0.7500 & 0.6250 \\
Feynman & 0.8947 & 0.7473 \\
Simple\_Test & 1.0000 & 0.2272 \\
\bottomrule
\end{tabular}
\end{table*}
\begin{table*}[ht]
\centering
\caption{Performance ($R^2$) across different number of variables in Ablation Experiments of Expression Separation.}
\label{tab:Separation_variable_R2}
\begin{small}
\begin{tabular}{ll ccccc}
\toprule
\multirow{2}{*}{Method} & \multirow{2}{*}{Variant} & \multicolumn{5}{c}{Variable Dimensionality} \\
\cmidrule(lr){3-7}
& & 2 & 3 & 4 & 5 & 6+ \\
\midrule
\multirow{3}{*}{E2E}
& Orig. & 0.9386 & 0.8129 & 0.6418 & 0.2491 & 0.0839 \\
& AIF. & 0.9618 & 0.9038 & 0.8313 & 0.6166 & 0.4820 \\
& DDRSR & 0.9785 & 0.9415 & 0.8833 & 0.8211 & 0.5801 \\
\midrule
\multirow{3}{*}{GP-GOMEA}
& Orig. & 0.9664 & 0.9440 & 0.8314 & 0.7703 & -0.4135 \\
& AIF. & 0.9729 & 0.9572 & 0.9023 & 0.8424 & -0.0152 \\
& DDRSR & 0.9880 & 0.9747 & 0.9593 & 0.9615 & 0.7573 \\
\midrule
\multirow{3}{*}{NGGP}
& Orig. & 0.9754 & 0.9036 & 0.6042 & 0.2963 & -0.0179 \\
& AIF. & 0.9883 & 0.9345 & 0.8432 & 0.6393 & 0.4866 \\
& DDRSR & 0.9988 & 0.9410 & 0.9102 & 0.7774 & 0.7500 \\
\midrule
\multirow{3}{*}{PySR}
& Orig. & 0.9889 & 0.9794 & 0.9463 & 0.7863 & 0.5602 \\
& AIF. & 0.9936 & 0.9872 & 0.9846 & 0.8605 & 0.4186 \\
& DDRSR & 0.9985 & 0.9989 & 0.9977 & 0.9762 & 0.8364 \\
\midrule
\multirow{3}{*}{MetaSymNet}
& Orig. & 0.8993 & 0.5268 & 0.2653 & 0.1255 & 0.0026 \\
& AIF. & 0.9638 & 0.7966 & 0.7423 & 0.4715 & 0.2261 \\
& DDRSR & 0.9819 & 0.8870 & 0.8428 & 0.6932 & 0.5285 \\
\bottomrule
\end{tabular}
\end{small}
\end{table*}
\clearpage
\begin{table*}[ht]
\centering
\caption{Performance ($R^2$) across different datasets in Ablation Experiments of Expression Separation.}
\label{tab:Separation_dataset_R2}
\begin{small}
\begin{tabular}{ll ccccc}
\toprule
Dataset & Variant & E2E & GP-GOMEA & NGGP & PySR & MetaSymNet \\
\midrule
\multirow{3}{*}{Keijzer} & Orig. & 0.9382 & 0.9896 & 0.9941 & 0.9946 & 0.8527 \\
 & AIF. & 0.9737 & 0.9910 & 0.9959 & 0.9837 & 0.9256 \\
 & DDRSR & 0.9889 & 0.9890 & 0.9998 & 0.9981 & 0.9360 \\
\cmidrule(lr){2-7}
\addlinespace[0.5em]
\multirow{3}{*}{Nguyen} & Orig. & 0.9507 & 0.9936 & 0.9935 & 0.9960 & 0.9822 \\
 & AIF. & 0.9580 & 0.9998 & 0.9997 & 0.9952 & 0.9864 \\
 & DDRSR & 0.9992 & 0.9999 & 0.9974 & 0.9988 & 0.9985 \\
\cmidrule(lr){2-7}
\addlinespace[0.5em]
\multirow{3}{*}{Constant} & Orig. & 0.9633 & 0.9978 & 0.9981 & 0.9212 & 0.9691 \\
\bottomrule
\end{tabular}
\end{small}
\end{table*}
\section{Detailed Experimental Results}
\begin{landscape}
\footnotesize
\setlength{\tabcolsep}{4pt}
\begin{longtable}{l c p{8cm} c p{8cm}}
\caption{Detailed Separation of DDRSR and AI Feynman in comprehensive experiments.} \label{tab:Comprehensive_separation} \\
\toprule
\textbf{Dataset} & \textbf{DDRSR Score} & \textbf{DDRSR Form} & \textbf{AIF Score} & \textbf{AIF Form} \\
\midrule
\endfirsthead
\toprule
\textbf{Dataset} & \textbf{DDRSR Score} & \textbf{DDRSR Form} & \textbf{AIF Score} & \textbf{AIF Form} \\
\midrule
\endhead
\bottomrule
\endlastfoot
Keijzer-10 & 1.0 & \texttt{exp(f1(x1)\allowbreak *\allowbreak f2(x2)\allowbreak )\allowbreak } & 0.0 & \texttt{f1(x1,x2)\allowbreak } \\
Keijzer-11 & 0.0 & \texttt{f1(x1,x2)\allowbreak } & 0.0 & \texttt{f1(x1,x2)\allowbreak } \\
Keijzer-12 & 1.0 & \texttt{f1(x1)\allowbreak +\allowbreak f2(x2)\allowbreak } & 1.0 & \texttt{f1(x1)\allowbreak  +\allowbreak  f2(x2)\allowbreak } \\
Keijzer-13 & 1.0 & \texttt{(f1(x1)\allowbreak )\allowbreak *\allowbreak (f2(x2)\allowbreak )\allowbreak } & 1.0 & \texttt{(f1(x1)\allowbreak )\allowbreak  *\allowbreak  (f2(x2)\allowbreak )\allowbreak } \\
Keijzer-14 & 0.0 & \texttt{f1(x1,x2)\allowbreak } & 0.0 & \texttt{f1(x1,x2)\allowbreak } \\
Keijzer-15 & 1.0 & \texttt{f1(x1)\allowbreak +\allowbreak f2(x2)\allowbreak } & 1.0 & \texttt{f1(x1)\allowbreak  +\allowbreak  f2(x2)\allowbreak } \\
Nguyen-9 & 1.0 & \texttt{f1(x1)\allowbreak +\allowbreak f2(x2)\allowbreak } & 1.0 & \texttt{f1(x1)\allowbreak  +\allowbreak  f2(x2)\allowbreak } \\
Nguyen-10 & 1.0 & \texttt{(f1(x1)\allowbreak )\allowbreak *\allowbreak (f2(x2)\allowbreak )\allowbreak } & 1.0 & \texttt{(f1(x1)\allowbreak )\allowbreak  *\allowbreak  (f2(x2)\allowbreak )\allowbreak } \\
\end{longtable}
\end{landscape}
\begin{landscape}
\footnotesize
\setlength{\tabcolsep}{4pt}
\begin{longtable}{l *{18}{c}}
\caption{Detailed Performance ($R^2$ and Recovery Rate) in comprehensive experiments.} \label{tab:Comprehensive_full_R2Rec} \\
\toprule
\multirow{3}{*}{\textbf{Dataset}} & \multicolumn{6}{c}{\textbf{GP-GOMEA}} & \multicolumn{6}{c}{\textbf{E2E}} & \multicolumn{6}{c}{\textbf{PySR}} \\
\cmidrule(lr){2-7} \cmidrule(lr){8-13} \cmidrule(lr){14-19}
& \multicolumn{2}{c}{Base} & \multicolumn{2}{c}{+AIF} & \multicolumn{2}{c}{+Div} & \multicolumn{2}{c}{Base} & \multicolumn{2}{c}{+AIF} & \multicolumn{2}{c}{+Div} & \multicolumn{2}{c}{Base} & \multicolumn{2}{c}{+AIF} & \multicolumn{2}{c}{+Div} \\
\cmidrule(lr){2-3} \cmidrule(lr){4-5} \cmidrule(lr){6-7} \cmidrule(lr){8-9} \cmidrule(lr){10-11} \cmidrule(lr){12-13} \cmidrule(lr){14-15} \cmidrule(lr){16-17} \cmidrule(lr){18-19}
& $R^2$ & Rec & $R^2$ & Rec & $R^2$ & Rec & $R^2$ & Rec & $R^2$ & Rec & $R^2$ & Rec & $R^2$ & Rec & $R^2$ & Rec & $R^2$ & Rec \\
\midrule
\endfirsthead
\multicolumn{19}{c}{{\tablename\ \thetable{} -- continued from previous page}} \\
\toprule
\multirow{3}{*}{\textbf{Dataset}} & \multicolumn{6}{c}{\textbf{GP-GOMEA}} & \multicolumn{6}{c}{\textbf{E2E}} & \multicolumn{6}{c}{\textbf{PySR}} \\
\cmidrule(lr){2-7} \cmidrule(lr){8-13} \cmidrule(lr){14-19}
& \multicolumn{2}{c}{Base} & \multicolumn{2}{c}{+AIF} & \multicolumn{2}{c}{+Div} & \multicolumn{2}{c}{Base} & \multicolumn{2}{c}{+AIF} & \multicolumn{2}{c}{+Div} & \multicolumn{2}{c}{Base} & \multicolumn{2}{c}{+AIF} & \multicolumn{2}{c}{+Div} \\
\cmidrule(lr){2-3} \cmidrule(lr){4-5} \cmidrule(lr){6-7} \cmidrule(lr){8-9} \cmidrule(lr){10-11} \cmidrule(lr){12-13} \cmidrule(lr){14-15} \cmidrule(lr){16-17} \cmidrule(lr){18-19}
& $R^2$ & Rec & $R^2$ & Rec & $R^2$ & Rec & $R^2$ & Rec & $R^2$ & Rec & $R^2$ & Rec & $R^2$ & Rec & $R^2$ & Rec & $R^2$ & Rec \\
\midrule
\endhead
\midrule
\multicolumn{19}{r}{{Continued on next page}} \\
\endfoot
\bottomrule
\endlastfoot
Keijzer-10 & 1.0000 & 1.00 & 1.0000 & 1.00 & 1.0000 & 1.00 & 0.7149 & 0.00 & 0.9800 & 0.00 & 0.9973 & 0.00 & 0.9987 & 0.00 & 0.9990 & 0.00 & 1.0000 & 1.00 \\
Keijzer-11 & 0.9848 & 0.00 & 0.9818 & 0.00 & 0.9828 & 0.00 & 0.9941 & 0.00 & 0.9923 & 0.00 & 0.9977 & 0.00 & 0.9904 & 0.00 & 1.0000 & 1.00 & 1.0000 & 1.00 \\
Keijzer-12 & 0.9611 & 0.00 & 1.0000 & 1.00 & 1.0000 & 1.00 & 0.9816 & 0.00 & 0.9939 & 0.00 & 0.9987 & 0.00 & 1.0000 & 1.00 & 1.0000 & 1.00 & 1.0000 & 1.00 \\
Keijzer-13 & 0.9992 & 0.00 & 1.0000 & 1.00 & 1.0000 & 1.00 & 0.9993 & 0.00 & 0.9981 & 0.00 & 0.9971 & 0.00 & 1.0000 & 1.00 & 1.0000 & 1.00 & 1.0000 & 1.00 \\
Keijzer-14 & 0.9930 & 0.00 & 0.9823 & 0.00 & 0.9867 & 0.00 & 0.9458 & 0.00 & 0.9340 & 0.00 & 0.9358 & 0.00 & 1.0000 & 1.00 & 1.0000 & 1.00 & 1.0000 & 1.00 \\
Keijzer-15 & 0.9999 & 0.80 & 1.0000 & 1.00 & 1.0000 & 1.00 & 0.9938 & 0.00 & 0.9997 & 0.00 & 0.9993 & 0.00 & 1.0000 & 1.00 & 1.0000 & 1.00 & 1.0000 & 1.00 \\
Nguyen-9 & 0.9968 & 0.00 & 1.0000 & 1.00 & 1.0000 & 1.00 & 0.9891 & 0.00 & 0.9984 & 0.00 & 0.9994 & 0.40 & 1.0000 & 1.00 & 1.0000 & 1.00 & 1.0000 & 1.00 \\
Nguyen-10 & 0.9988 & 0.00 & 0.9999 & 0.80 & 1.0000 & 1.00 & 0.9992 & 0.00 & 0.9988 & 0.00 & 0.9997 & 0.00 & 1.0000 & 1.00 & 1.0000 & 1.00 & 1.0000 & 1.00 \\
\end{longtable}
\end{landscape}

\begin{landscape}
\footnotesize
\setlength{\tabcolsep}{4pt}
\begin{longtable}{l c p{8cm} c p{8cm}}
\caption{Detailed Separation of DDRSR and AI Feynman in Ablation Experiments of Expression Separation.} \label{tab:Separation_separation} \\
\toprule
\textbf{Dataset} & \textbf{DDRSR Score} & \textbf{DDRSR Form} & \textbf{AIF Score} & \textbf{AIF Form} \\
\midrule
\endfirsthead
\multicolumn{5}{c}{{\tablename\ \thetable{} -- continued from previous page}} \\
\toprule
\textbf{Dataset} & \textbf{DDRSR Score} & \textbf{DDRSR Form} & \textbf{AIF Score} & \textbf{AIF Form} \\
\midrule
\endhead
\midrule
\multicolumn{5}{r}{{Continued on next page}} \\
\endfoot
\bottomrule
\endlastfoot
Keijzer-10 & 1.0 & exp(f1(x1)\allowbreak *\allowbreak f2(x2)\allowbreak )\allowbreak  & 0.0 & f1(x1,x2)\allowbreak  \\
Keijzer-11 & 0.0 & f1(x1,x2)\allowbreak  & 0.0 & f1(x1,x2)\allowbreak  \\
Keijzer-12 & 1.0 & f1(x1)\allowbreak +\allowbreak f2(x2)\allowbreak  & 1.0 & f1(x1)\allowbreak +\allowbreak f2(x2)\allowbreak  \\
Keijzer-13 & 1.0 & (f1(x1)\allowbreak )\allowbreak *\allowbreak (f2(x2)\allowbreak )\allowbreak  & 1.0 & (f1(x1)\allowbreak )\allowbreak *\allowbreak (f2(x2)\allowbreak )\allowbreak  \\
Keijzer-14 & 0.0 & f1(x1,x2)\allowbreak  & 0.0 & f1(x1,x2)\allowbreak  \\
Keijzer-15 & 1.0 & f1(x1)\allowbreak +\allowbreak f2(x2)\allowbreak  & 1.0 & f1(x1)\allowbreak +\allowbreak f2(x2)\allowbreak  \\
Nguyen-9 & 1.0 & f1(x1)\allowbreak +\allowbreak f2(x2)\allowbreak  & 1.0 & f1(x1)\allowbreak +\allowbreak f2(x2)\allowbreak  \\
Nguyen-10 & 1.0 & (f1(x1)\allowbreak )\allowbreak *\allowbreak (f2(x2)\allowbreak )\allowbreak  & 1.0 & (f1(x1)\allowbreak )\allowbreak *\allowbreak (f2(x2)\allowbreak )\allowbreak  \\
\end{longtable}
\end{landscape}
\begin{landscape}
\footnotesize
\setlength{\tabcolsep}{3.5pt}
\begin{longtable}{l *{15}{c}}
\caption{Detailed Performance ($R^2$) in Ablation Experiments of Expression Separation.} \label{tab:Separation_full_R2} \\
\toprule
\textbf{Dataset} & \multicolumn{3}{c}{\textbf{E2E}} & \multicolumn{3}{c}{\textbf{GP-GOMEA}} & \multicolumn{3}{c}{\textbf{NGGP}} & \multicolumn{3}{c}{\textbf{PySR}} & \multicolumn{3}{c}{\textbf{MetaSymNet}} \\
\cmidrule(lr){2-4} \cmidrule(lr){5-7} \cmidrule(lr){8-10} \cmidrule(lr){11-13} \cmidrule(lr){14-16}
& Base & +AIF & +Div & Base & +AIF & +Div & Base & +AIF & +Div & Base & +AIF & +Div & Base & +AIF & +Div \\
\midrule
\endfirsthead
\multicolumn{16}{c}{{\tablename\ \thetable{} -- continued from previous page}} \\
\toprule
\textbf{Dataset} & \multicolumn{3}{c}{\textbf{E2E}} & \multicolumn{3}{c}{\textbf{GP-GOMEA}} & \multicolumn{3}{c}{\textbf{NGGP}} & \multicolumn{3}{c}{\textbf{PySR}} & \multicolumn{3}{c}{\textbf{MetaSymNet}} \\
\cmidrule(lr){2-4} \cmidrule(lr){5-7} \cmidrule(lr){8-10} \cmidrule(lr){11-13} \cmidrule(lr){14-16}
& Base & +AIF & +Div & Base & +AIF & +Div & Base & +AIF & +Div & Base & +AIF & +Div & Base & +AIF & +Div \\
\midrule
\endhead
\midrule
\multicolumn{16}{r}{{Continued on next page}} \\
\endfoot
\bottomrule
\endlastfoot
Keijzer-10 & 0.7149 & 0.9360 & 0.9991 & 1.0000 & 1.0000 & 1.0000 & 0.9948 & 0.9998 & 1.0000 & 0.9784 & 0.9995 & 1.0000 & 0.9524 & 0.9562 & 0.9972 \\
Keijzer-11 & 0.9941 & 0.9961 & 0.9937 & 0.9848 & 0.9608 & 0.9415 & 0.9958 & 0.9785 & 0.9992 & 0.9948 & 0.9927 & 0.9962 & 0.2511 & 0.6666 & 0.6403 \\
Keijzer-12 & 0.9816 & 0.9783 & 0.9999 & 0.9611 & 0.9998 & 1.0000 & 0.9830 & 1.0000 & 1.0000 & 0.9960 & 0.9980 & 0.9949 & 0.9767 & 0.9999 & 0.9889 \\
Keijzer-13 & 0.9993 & 0.9997 & 0.9998 & 0.9992 & 1.0000 & 0.9996 & 0.9997 & 1.0000 & 1.0000 & 0.9999 & 1.0000 & 1.0000 & 0.9969 & 0.9998 & 1.0000 \\
Keijzer-14 & 0.9458 & 0.9327 & 0.9413 & 0.9930 & 0.9856 & 0.9934 & 0.9954 & 0.9973 & 0.9996 & 0.9997 & 0.9141 & 0.9996 & 0.9758 & 0.9311 & 0.9897 \\
Keijzer-15 & 0.9938 & 0.9994 & 0.9997 & 0.9999 & 1.0000 & 1.0000 & 0.9963 & 1.0000 & 1.0000 & 0.9988 & 0.9980 & 0.9982 & 0.9636 & 1.0000 & 1.0000 \\
Nguyen-9 & 0.9891 & 0.9972 & 0.9999 & 0.9968 & 0.9999 & 1.0000 & 0.9997 & 0.9997 & 1.0000 & 1.0000 & 0.9973 & 1.0000 & 0.9891 & 1.0000 & 1.0000 \\
Nguyen-10 & 0.9992 & 0.9998 & 0.9994 & 0.9988 & 1.0000 & 1.0000 & 0.9998 & 1.0000 & 0.9873 & 1.0000 & 0.9998 & 0.9999 & 0.9836 & 0.9782 & 0.9992 \\
\end{longtable}
\end{landscape}
\begin{landscape}
\footnotesize
\setlength{\tabcolsep}{4pt}
\begin{longtable}{l c p{8cm} c p{8cm}}
\caption{Detailed Separation of DDRSR and AI Feynman in Ablation Experiments of Variable Composition} \label{tab:Translational_Symmetry_separation} \\
\toprule
\textbf{Dataset} & \textbf{DDRSR Score} & \textbf{DDRSR Form} & \textbf{AIF Score} & \textbf{AIF Form} \\
\midrule
\endfirsthead
\toprule
\textbf{Dataset} & \textbf{DDRSR Score} & \textbf{DDRSR Form} & \textbf{AIF Score} & \textbf{AIF Form} \\
\midrule
\endhead
\bottomrule
\endlastfoot
complex-test1 & 0.0 & \texttt{f1(x1,x2,x3,x4,x5)\allowbreak } & 0.0 & \texttt{f1(x1,x2,x3,x4,x5)\allowbreak } \\
complex-test2 & 0.0 & \texttt{f1(x1,x2,x3,x4,x5)\allowbreak } & 0.0 & \texttt{f1(x1,x2,x3,x4,x5)\allowbreak } \\
complex-test3 & 1.0 & \texttt{f1((x1\textasciicircum{}(1)\allowbreak *\allowbreak x2\textasciicircum{}(2)\allowbreak *\allowbreak x5\textasciicircum{}(-1)\allowbreak )\allowbreak ,x3,x4)\allowbreak } & 0.5 & \texttt{f1(x1/x5,x2,x3,x4)\allowbreak } \\
complex-test4 & 1.0 & \texttt{f1((x1+\allowbreak (1.2000)\allowbreak *\allowbreak x2)\allowbreak ,x3,(x4+\allowbreak (1.1667)\allowbreak *\allowbreak x5)\allowbreak )\allowbreak } & 0.0 & \texttt{f1(x1,x2,x3,x4,x5)\allowbreak } \\
complex-test5 & 1.0 & \texttt{f1((x1+\allowbreak (2.0000)\allowbreak *\allowbreak x2+\allowbreak (3.0000)\allowbreak *\allowbreak x3)\allowbreak ,(x4\textasciicircum{}(1)\allowbreak *\allowbreak x5\textasciicircum{}(2)\allowbreak )\allowbreak )\allowbreak } & 0.0 & \texttt{f1(x1,x2,x3,x4,x5)\allowbreak } \\
complex-test6 & 1.0 & \texttt{f1((x1\textasciicircum{}(1)\allowbreak *\allowbreak x2\textasciicircum{}(2)\allowbreak )\allowbreak ,(x3+\allowbreak (1.5000)\allowbreak *\allowbreak x4)\allowbreak ,(x5\textasciicircum{}(2)\allowbreak *\allowbreak x6\textasciicircum{}(3)\allowbreak )\allowbreak )\allowbreak } & 0.0 & \texttt{f1(x1,x2,x3,x4,x5,x6)\allowbreak } \\
complex-test7 & 1.0 & \texttt{f1(x1,x2,(x3+\allowbreak (2.0000)\allowbreak *\allowbreak x4)\allowbreak ,((x5+\allowbreak (-0.5000)\allowbreak *\allowbreak x6)\allowbreak \textasciicircum{}(1)\allowbreak *\allowbreak x7\textasciicircum{}(-1)\allowbreak *\allowbreak x8\textasciicircum{}(-2)\allowbreak )\allowbreak )\allowbreak } & 0.0 & \texttt{f1(x1,x2,x3,x4,x5,x6,x7,x8)\allowbreak } \\
\end{longtable}
\end{landscape}
\begin{landscape}
\footnotesize
\setlength{\tabcolsep}{3.5pt}
\begin{longtable}{l *{15}{c}}
\caption{Detailed Performance ($R^2$) in Ablation Experiments of Variable Composition.} \label{tab:Translational_Symmetry_full_R2} \\
\toprule
\textbf{Dataset} & \multicolumn{3}{c}{\textbf{E2E}} & \multicolumn{3}{c}{\textbf{GP-GOMEA}} & \multicolumn{3}{c}{\textbf{NGGP}} & \multicolumn{3}{c}{\textbf{PySR}} & \multicolumn{3}{c}{\textbf{MetaSymNet}} \\
\cmidrule(lr){2-4} \cmidrule(lr){5-7} \cmidrule(lr){8-10} \cmidrule(lr){11-13} \cmidrule(lr){14-16}
& Base & +AIF & +Div & Base & +AIF & +Div & Base & +AIF & +Div & Base & +AIF & +Div & Base & +AIF & +Div \\
\midrule
\endfirsthead
\multicolumn{16}{c}{{\tablename\ \thetable{} -- continued from previous page}} \\
\toprule
\textbf{Dataset} & \multicolumn{3}{c}{\textbf{E2E}} & \multicolumn{3}{c}{\textbf{GP-GOMEA}} & \multicolumn{3}{c}{\textbf{NGGP}} & \multicolumn{3}{c}{\textbf{PySR}} & \multicolumn{3}{c}{\textbf{MetaSymNet}} \\
\cmidrule(lr){2-4} \cmidrule(lr){5-7} \cmidrule(lr){8-10} \cmidrule(lr){11-13} \cmidrule(lr){14-16}
& Base & +AIF & +Div & Base & +AIF & +Div & Base & +AIF & +Div & Base & +AIF & +Div & Base & +AIF & +Div \\
\midrule
\endhead
\midrule
\multicolumn{16}{r}{{Continued on next page}} \\
\endfoot
\bottomrule
\endlastfoot
complex-test1 & 0.7746 & 0.7487 & 0.7124 & 0.9938 & 0.9999 & 0.9999 & 0.6533 & 0.7181 & 0.6723 & 0.9908 & 0.9927 & 0.9821 & 0.6264 & 0.6707 & 0.5972 \\
complex-test2 & 0.7398 & 0.8061 & 0.8867 & 0.9723 & 0.9769 & 0.9747 & 0.1501 & 0.1622 & 0.2232 & 0.8969 & 0.9345 & 0.9291 & 0.6514 & 0.6671 & 0.6844 \\
complex-test3 & 0.2694 & 0.9337 & 0.9997 & 0.9497 & 0.8824 & 0.9946 & -0.9234 & 0.9445 & 0.9950 & 0.9960 & 0.9930 & 1.0000 & -0.0992 & 0.0642 & 0.1164 \\
complex-test4 & 0.9681 & 0.9711 & 0.9985 & 0.9534 & 0.9877 & 0.9871 & 0.8541 & 0.7823 & 0.9532 & 0.9992 & 0.9941 & 0.9999 & 0.6522 & 0.8087 & 0.9726 \\
complex-test5 & 0.8622 & 0.7616 & 0.9997 & 0.9568 & 0.9183 & 0.9832 & 0.7094 & 0.7875 & 0.2109 & 0.9998 & 0.9997 & 1.0000 & 0.7743 & 0.8433 & 0.9844 \\
complex-test6 & 0.8606 & 0.6253 & 0.9558 & 0.7023 & 0.7561 & 1.0000 & -0.5006 & -0.8065 & 0.9782 & 0.9925 & 0.9790 & 0.9984 & -0.0356 & 0.3749 & 0.8935 \\
complex-test7 & -0.7962 & -0.3475 & 0.9364 & 0.2409 & -0.5462 & 0.8854 & -0.4746 & -0.0007 & 0.9166 & 0.0231 & -0.0211 & 0.9167 & -0.3776 & -0.5273 & 0.4535 \\
complex-test8 & 0.1992 & -0.0438 & 0.2381 & 0.1614 & 0.2692 & 0.3756 & -0.0297 & -0.4877 & 0.1147 & 0.1343 & 0.5240 & 0.6964 & 0.0229 & -0.1689 & -0.4380 \\
\end{longtable}
\end{landscape}

\end{document}